\documentclass{article}

\usepackage{microtype}
\usepackage{graphicx}
\usepackage{subfigure}
\usepackage{booktabs} % for professional tables

\usepackage{hyperref}

\usepackage[accepted]{icml2018}

\icmltitlerunning{Generalized Robust Bayesian Committee Machine for Large-scale Gaussian Process Regression}

\usepackage{amsmath,amssymb,bm,color}

\usepackage{amsthm}
\theoremstyle{remark}
\newtheorem{rem}{Remark}

\theoremstyle{plain}
\newtheorem{thm}{Theorem}
\newtheorem{prop}[thm]{Proposition}

\theoremstyle{definition}
\newtheorem{defn}{Definition}

\begin{document}
	
	\twocolumn[
	\icmltitle{Generalized Robust Bayesian Committee Machine for Large-scale Gaussian Process Regression}
	
	% It is OKAY to include author information, even for blind
	% submissions: the style file will automatically remove it for you
	% unless you've provided the [accepted] option to the icml2018
	% package.
	
	% List of affiliations: The first argument should be a (short)
	% identifier you will use later to specify author affiliations
	% Academic affiliations should list Department, University, City, Region, Country
	% Industry affiliations should list Company, City, Region, Country
	
	% You can specify symbols, otherwise they are numbered in order.
	% Ideally, you should not use this facility. Affiliations will be numbered
	% in order of appearance and this is the preferred way.
	\icmlsetsymbol{equal}{*}
	
	\begin{icmlauthorlist}
		\icmlauthor{Haitao Liu}{RRNTU}
		\icmlauthor{Jianfei Cai}{NTU}
		\icmlauthor{Yi Wang}{RR}
		\icmlauthor{Yew-Soon Ong}{NTU,NTU2}
	\end{icmlauthorlist}
	
	\icmlaffiliation{RRNTU}{Rolls-Royce@NTU Corporate Lab, Nanyang Technological University, Singapore 637460}
	\icmlaffiliation{NTU}{School of Computer Science and Engineering, Nanyang Technological University, Singapore 639798}
	\icmlaffiliation{NTU2}{Data Science and Artificial Intelligence Research Center, Nanyang Technological University, Singapore 639798}
	\icmlaffiliation{RR}{Applied Technology Group, Rolls-Royce Singapore, 6 Seletar Aerospace Rise, Singapore 797575}
	
	\icmlcorrespondingauthor{Haitao Liu}{htliu@ntu.edu.sg}
	
	% You may provide any keywords that you
	% find helpful for describing your paper; these are used to populate
	% the "keywords" metadata in the PDF but will not be shown in the document
	\icmlkeywords{Machine Learning, ICML}
	
	\vskip 0.3in
	]
	
	% this must go after the closing bracket ] following \twocolumn[ ...
	
	% This command actually creates the footnote in the first column
	% listing the affiliations and the copyright notice.
	% The command takes one argument, which is text to display at the start of the footnote.
	% The \icmlEqualContribution command is standard text for equal contribution.
	% Remove it (just {}) if you do not need this facility.
	
	\printAffiliationsAndNotice{}  % leave blank if no need to mention equal contribution
	%\printAffiliationsAndNotice{\icmlEqualContribution} % otherwise use the standard text.
	
	\begin{abstract}
		In order to scale standard Gaussian process (GP) regression to large-scale datasets, aggregation models employ factorized training process and then combine predictions from distributed experts. The state-of-the-art aggregation models, however, either provide inconsistent predictions or require time-consuming aggregation process. We first prove the inconsistency of typical aggregations using disjoint or random data partition, and then present a consistent yet efficient aggregation model for large-scale GP. The proposed model inherits the advantages of aggregations, e.g., closed-form inference and aggregation, parallelization and distributed computing. Furthermore, theoretical and empirical analyses reveal that the new aggregation model performs better due to the consistent predictions that converge to the true underlying function when the training size approaches infinity.
	\end{abstract}
	
	\section{Introduction}
	Gaussian process (GP) \cite{rasmussen2006gaussian} is a well-known statistical learning model extensively used in various scenarios, e.g., regression, classification, optimization \cite{shahriari2016taking}, visualization \cite{lawrence2005probabilistic}, active learning \cite{fu2013survey, liu2017adaptive} and multi-task learning \cite{alvarez2012kernels, liu2018remarks}. Given the training set $\bm{X} = \{\bm{x}_i \in R^d \}_{i=1}^n$ and the observation set $\bm{y} = \{ y(\bm{x}_i) \in R \}_{i=1}^n$, as an approximation of the underlying function $\eta: R^d \rightarrow R$, GP provides informative predictive distributions at test points.
	
	However, the most prominent weakness of the full GP is that it scales poorly with the training size. Given $n$ data points, the time complexity of a standard GP paradigm scales as $\mathcal{O}(n^3)$ in the training process due to the inversion of an $n \times n$ covariance matrix; it scales as $\mathcal{O}(n^2)$ in the prediction process due to the matrix-vector operation. This weakness confines the full GP to training data of size $\mathcal{O}(10^4)$. 
	
	To cope with large-scale regression, various computationally efficient approximations have been presented. The sparse approximations reviewed in \cite{quinonero2005unifying} employ $m$ ($m \ll n$) inducing points to summarize the whole training data \cite{seeger2003fast, snelson2006sparse, snelson2007local, titsias2009variational, bauer2016understanding}, thus reducing the training complexity of full GP to $\mathcal{O}(nm^2)$ and the predicting complexity to $\mathcal{O}(nm)$. The complexity can be further reduced through distributed inference, stochastic variational inference or Kronecker structure \cite{hensman2013gaussian, gal2014distributed, wilson2015kernel, hoang2016distributed, peng2017asynchronous}. A main drawback of sparse approximations, however, is that the representational capability is limited by the number of inducing points \cite{moore2015gaussian}. For example, for a quick-varying function, the sparse approximations need many inducing points to capture the local structures. That is, this kind of scheme has not reduced the scaling of the complexity \cite{bui2014tree}.
	
	The method exploited in this article belongs to the aggregation models \cite{hinton2002training, tresp2000bayesian, cao2014generalized, deisenroth2015distributed, rulliere2017nested}, also known as consensus statistical methods \cite{genest1986combining, ranjan2010combining}. This kind of scheme produces the final predictions by the aggregation of $M$ sub-models (GP experts) respectively trained on the subsets $\{\mathcal{D}_i=\{\bm{X}_i,\bm{y}_i \} \}_{i=1}^M$ of $\mathcal{D}=\{\bm{X},\bm{y}\}$, thus distributing the computations to ``local'' experts. Particularly, due to the product of experts, the aggregation scheme derives a factorized marginal likelihood for efficient training; and then it combines the experts' posterior distributions according to a certain aggregation criterion. In comparison to sparse approximations, the aggregation models (i) operate directly on the full training data, (ii) require no additional inducing or variational parameters and (iii) distribute the computations on individual experts for straightforward parallelization \cite{tavassolipour2017learning}, thus scaling them to arbitrarily large training data. In comparison to typical local GPs \cite{snelson2007local,  park2011domain}, the aggregations smooth out the ugly discontinuity by the product of posterior distributions from GP experts. Note that the aggregation methods are different from the mixture-of-experts \cite{rasmussen2002infinite, yuan2009variational}, which suffers from intractable inference and is mainly developed for non-stationary regression.
	
	However, it has been pointed out \cite{rulliere2017nested} that there exists a particular type of training data such that typical aggregations, e.g., product-of-experts (PoE) \cite{hinton2002training, cao2014generalized} and Bayesian committee machine (BCM) \cite{tresp2000bayesian, deisenroth2015distributed}, cannot offer \textit{consistent} predictions, where ``\textit{consistent}'' means the aggregated predictive distribution can converge to the true underlying predictive distribution when the training size $n$ approaches infinity.
	
	The major contributions of this paper are three-fold. We first prove the inconsistency of typical aggregation models, e.g., the overconfident or conservative prediction variances illustrated in Fig.~\ref{Fig_Toy_comparison}, using conventional disjoint or random data partition. Thereafter, we present a consistent yet efficient aggregation model for large-scale GP regression. Particularly, the proposed generalized robust Bayesian committee machine (GRBCM) selects a global subset to communicate with the remaining subsets, leading to the consistent aggregated predictive distribution derived under the Bayes rule. Finally, theoretical and empirical analyses reveal that GRBCM outperforms existing aggregations due to the consistent yet efficient predictions. We release the demo codes in \url{https://github.com/LiuHaiTao01/GRBCM}.

	\section{Aggregation models revisited} \label{Sec_2}
	\subsection{Factorized training} \label{Sec_2.1}
	A GP usually places a probability distribution over the latent function space as $f(\bm{x}) \sim \mathcal{GP}(0, k(\bm{x},\bm{x}'))$, which is defined by the zero mean and the covariance $k(\bm{x},\bm{x}')$. The well-known squared exponential (SE) covariance function is
	\begin{equation} \label{Eq_SE_Kernel}
	k(\bm{x},\bm{x}')= \sigma^2_{f} \exp \left(- \frac{1}{2} \sum_{i=1}^d \frac{(x_i-x'_i)^2}{l_i^2} \right),
	\end{equation}
	where $\sigma^2_{f}$ is an output scale amplitude, and $l_i$ is an input length-scale along the $i$th dimension. Given the noisy observation $y(\bm{x}) = f(\bm{x}) + \epsilon$ where the $i.i.d.$ noise follows $\epsilon \sim \mathcal{N}(0, \sigma^2_{\epsilon})$ and the training data $\mathcal{D}$, we have the marginal likelihood $p(\bm{y}|\bm{X},\bm{\theta}) = \mathcal{N}(\mathbf{0}, k(\bm{X}, \bm{X}) + \sigma^2_{\epsilon} \bm{I})$ where $\bm{\theta}$ represents the hyperparameters to be inferred. 
	
	In order to train the GP on large-scale datasets, the aggregation models introduce a factorized training process. It first partitions the training set $\mathcal{D}$ into $M$ subsets $\mathcal{D}_i = \{\bm{X}_i, \bm{y}_i \}$, $1 \le i \le M$, and then trains GP on $\mathcal{D}_i$ as an expert $\mathcal{M}_i$. In data partition, we can assign the data points randomly to the experts (\textit{random partition}), or assign disjoint subsets obtained by clustering techniques to the experts (\textit{disjoint partition}). Ignoring the correlation between the experts $\{\mathcal{M}_i\}_{i=1}^M$ leads to the factorized approximation as
	\begin{equation} \label{Eq_approxML}
	p(\bm{y}|\bm{X},\bm{\theta}) \approx \prod_{i=1}^M p_i(\bm{y}_i|\bm{X}_i,\bm{\theta}_i),
	\end{equation}
	where $p_i(\bm{y}_i|\bm{X}_i,\bm{\theta}_i) \sim \mathcal{N}(\mathbf{0}, \bm{K}_i + \sigma^2_{\epsilon, i} \bm{I}_i)$ with $\bm{K}_i = k(\bm{X}_i, \bm{X}_i) \in R^{n_i \times n_i}$ and $n_i$ being the training size of $\mathcal{M}_i$. Note that for simplicity all the $M$ GP experts in \eqref{Eq_approxML} share the same hyperparameters as $\bm{\theta}_i = \bm{\theta}$  \cite{deisenroth2015distributed}. The factorization \eqref{Eq_approxML} degenerates the full covariance matrix $\bm{K} = k(\bm{X},\bm{X})$ into a diagonal block matrix $\mathrm{diag}[\bm{K}_1, \cdots, \bm{K}_M]$, leading to $\bm{K}^{-1} \approx \mathrm{diag}[\bm{K}_1^{-1}, \cdots, \bm{K}_M^{-1}]$. Hence, compared to the full GP, the complexity of the factorized training process is reduced to $\mathcal{O}(nm_0^2)$ given $n_i = m_0 = n/M$, $1 \le i \le M$.
	
	Conditioned on the related subset $\mathcal{D}_i$, the predictive distribution $p_i(y_*|\mathcal{D}_i, \bm{x}_*) \sim \mathcal{N}(\mu_i(\bm{x}_*), \sigma^2_i(\bm{x}_*))$ of $\mathcal{M}_i$ has\footnote{Instead of using $p_i(f_*|\mathcal{D}_i, \bm{x}_*)$ in \cite{deisenroth2015distributed}, we here consider the aggregations in a general scenario where each expert has all its belongings at hand.}
	\begin{subequations}
		\begin{align}
		\mu_i(\bm{x}_*) &= {\bm{k}}_{i*}^\mathsf{T}[\bm{K}_i+\sigma_{\epsilon}^2 \bm{I}]^{-1}\bm{y}_i, \label{Eq_mu_Mi} \\
		\sigma_i^2(\bm{x}_*) &= k({\bm{x}}_*,{\bm{x}}_*) -  {\bm{k}}_{i*}^\mathsf{T}[\bm{K}_i+\sigma_{\epsilon}^2 \bm{I}]^{-1} {\bm{k}}_{i*} + \sigma^2_{\epsilon},
		\end{align}
	\end{subequations}
	where $\bm{k}_{i*} = k(\bm{X}_i, \bm{x}_*)$. Thereafter, the experts' predictions $\{\mu_i, \sigma_i^2\}_{i=1}^M$ are combined by the following aggregation methods to perform the final predicting.

	\subsection{Prediction aggregation}
	The state-of-the-art aggregation methods include PoE \cite{hinton2002training, cao2014generalized}, BCM \cite{tresp2000bayesian, deisenroth2015distributed}, and nested pointwise aggregation of experts (NPAE) \cite{rulliere2017nested}. 
	
	For the PoE and BCM family, the aggregated prediction mean and precision are generally formulated as
	\begin{subequations}
		\begin{align}
		\mu_{\mathcal{A}}(\bm{x}_*) &= \sigma_{\mathcal{A}}^2(\bm{x}_*) \sum_{i=1}^M \beta_i \sigma_i^{-2}(\bm{x}_*) \mu_i(\bm{x}_*), \label{Eq_mu_general} \\
		\sigma_{\mathcal{A}}^{-2}(\bm{x}_*) &= \sum_{i=1}^M \beta_i \sigma_i^{-2}(\bm{x}_*) + (1-\sum_{i=1}^M \beta_i)\sigma_{**}^{-2}, \label{Eq_s2_general}
		\end{align}
	\end{subequations}
	where the prior variance $\sigma_{**}^2 = k(\bm{x}_*,\bm{x}_*) + \sigma_{\epsilon}^2$, which is a correction term to $\sigma^{-2}_{\mathcal{A}}$, is only available for the BCM family; and $\beta_i$ is the weight of the expert $\mathcal{M}_i$ at $\bm{x}_*$.
	
	The predictions of the PoE family, which omit the prior precision $\sigma^{-2}_{**}$ in \eqref{Eq_s2_general}, are derived from the product of $M$ experts as
	\begin{equation} \label{Eq_predict_product}
	p_{\mathcal{A}}(y_*|\mathcal{D}, \bm{x}_*) = \prod_{i=1}^M p^{\beta_i}_i(y_*|\mathcal{D}_i, \bm{x}_*).
	\end{equation}
	The original PoE \cite{hinton2002training} employs the constant weight $\beta_i = 1$, resulting in the aggregated prediction variances that vanish with increasing $M$. On the contrary, the generalized PoE (GPoE) \cite{cao2014generalized} considers a varying $\beta_i = 0.5(\log\sigma_{**}^{2} - \log\sigma_i^2(\bm{x}_*))$, which represents the difference in the differential entropy between the prior $p(y_*|\bm{x}_*)$ and the posterior $p(y_*|\mathcal{D}_i, \bm{x}_*)$,  to weigh the contribution of $\mathcal{M}_i$ at $\bm{x}_*$. This varying $\beta_i$ brings the flexibility of increasing or reducing the importance of experts based on the predictive uncertainty. However, the varying $\beta_i$ may produce undesirable errors for GPoE. For instance, when $\bm{x}_*$ is far away from the training data such that $\sigma_i^2(\bm{x}_*) \to \sigma_{**}^{2}$, we have $\beta_i \to 0$ and $\sigma^{2}_{\mathrm{GPoE}} \to \infty$.
	
	The BCM family, which is opposite to the PoE family, explicitly incorporates the GP prior $p(y_*|\bm{x}_*)$ when combining predictions. For two experts $\mathcal{M}_i$ and $\mathcal{M}_j$, BCM introduces a conditional independence assumption $\mathcal{D}_i \perp \mathcal{D}_j|y_*$, leading to the aggregated predictive distribution as
	\begin{equation}
	p_{\mathcal{A}}(y_*|\mathcal{D}, \bm{x}_*) = \frac{\prod_{i=1}^M p_i^{\beta_i}(y_*|\mathcal{D}_i, \bm{x}_*)}{p^{\sum_i \beta_i-1}(y_*|\bm{x}_*)}.
	\end{equation}
	The original BCM \cite{tresp2000bayesian} employs $\beta_i = 1$ but its predictions suffer from weak experts when leaving the data. Hence, inspired by GPoE, the robust BCM (RBCM) \cite{deisenroth2015distributed} uses a varying $\beta_i$ to produce robust predictions by reducing the weights of weak experts. When $\bm{x}_*$ is far away from the training data $\bm{X}$, the correction term brought by the GP prior in \eqref{Eq_s2_general} helps the (R)BCM's prediction variance recover $\sigma^2_{**}$. However, given $M=1$, the predictions of RBCM as well as GPoE cannot recover the full GP predictions because usually $\beta_1 = 0.5(\log\sigma_{**}^{2} - \log\sigma_1^2(\bm{x}_*)) = 0.5(\log\sigma_{**}^{2} - \log\sigma_{full}^2(\bm{x}_*)) \ne 1$.
	
	To achieve computation gains, the above aggregations introduce additional independence assumption for the experts' predictions, which however is often violated in practice and yields poor results. Hence, in the aggregation process, NPAE \cite{rulliere2017nested} regards the prediction mean $\mu_i(\bm{x}_*)$ in \eqref{Eq_mu_Mi} as a random variable by assuming that $\bm{y}_i$ has not yet been observed, thus allowing for considering the covariances between the experts' predictions. Thereafter, for the random vector $[\mu_1, \cdots, \mu_M, y_*]^{\mathsf{T}}$, the covariances are derived as
	\begin{subequations}
		\begin{align}
		\mathrm{cov}[\mu_i,y_*] & = {\bm{k}}_{i*}^\mathsf{T} \bm{K}_{i, \epsilon}^{-1} {\bm{k}}_{i*}, \\
		\mathrm{cov}[\mu_i,\mu_j] & = \left\lbrace \begin{aligned}
		&{\bm{k}}_{i*}^\mathsf{T} \bm{K}_{i,\epsilon}^{-1} \bm{K}_{ij} \bm{K}_{j,\epsilon}^{-1}{\bm{k}}_{j*}, &i\ne j , \\
		&{\bm{k}}_{i*}^\mathsf{T} \bm{K}_{i,\epsilon}^{-1} \bm{K}_{ij,\epsilon} \bm{K}_{j,\epsilon}^{-1}{\bm{k}}_{j*}, &i =j,
		\end{aligned}
		\right.
		\end{align}
	\end{subequations}
	where $\bm{K}_{ij} = k(\bm{X}_i, \bm{X}_j) \in R^{n_i \times n_j}$, $\bm{K}_{i,\epsilon} = \bm{K}_i + \sigma^2_{\epsilon}\bm{I}$, $\bm{K}_{j,\epsilon} = \bm{K}_j + \sigma^2_{\epsilon}\bm{I}$, and $\bm{K}_{ij,\epsilon} = \bm{K}_{ij} + \sigma^2_{\epsilon}\bm{I}$. With these covariances, a nested GP training process is performed to derive the aggregated prediction mean and variance as
	\begin{subequations}
		\begin{align}
		\mu_{\mathrm{NPAE}}(\bm{x}_*) &= \bm{k}_{\mathcal{A}*}^{\mathsf{T}}\bm{K}_{\mathcal{A}}^{-1}\bm{\mu}, \label{Eq_mu_NPAE}\\
		\sigma^2_{\mathrm{NPAE}}(\bm{x}_*) &= k({\bm{x}}_*,{\bm{x}}_*) - \bm{k}_{\mathcal{A}*}^{\mathsf{T}}\bm{K}_{\mathcal{A}}^{-1}\bm{k}_{\mathcal{A}*} + \sigma^2_{\epsilon}, \label{Eq_s2_NPAE}
		\end{align}
	\end{subequations}
	where $\bm{k}_{\mathcal{A}*} \in R^{M \times 1}$ has the $i$th element as $\mathrm{cov}[\mu_i,y_*]$, $\bm{K}_{\mathcal{A}} \in R^{M \times M}$ has $\bm{K}_{\mathcal{A}}^{ij} = \mathrm{cov}[\mu_i,\mu_j]$, and $\bm{\mu} = [\mu_1(\bm{x}_*),\cdots,\mu_M(\bm{x}_*)]^{\mathsf{T}}$. The NPAE is capable of providing \textit{consistent} predictions at the cost of implementing a much more time-consuming aggregation because of the inversion of $\bm{K}_{\mathcal{A}}$ at each test point.

	\subsection{Discussions of existing aggregations}
	Though showcasing promising results \cite{deisenroth2015distributed}, given that $n \to \infty$ and the experts are noise-free GPs, (G)PoE and (R)BCM have been proved to be \textit{inconsistent}, since  there exists particular triangular array of data points that are dense in the input domain $\Omega$ such that the prediction variances do not go to zero \cite{rulliere2017nested}.
	
	Particularly, we further show below the inconsistency of (G)PoE and (R)BCM using two typical data partitions (random and disjoint partition) in the scenario where the observations are blurred with noise. Note that since GPoE using a varying $\beta_i$ may produce undesirable errors, we adopt $\beta_i = 1/M$ as suggested in \cite{deisenroth2015distributed}. Now the GPoE's prediction mean is the same as that of PoE; but the prediction variance blows up as $M$ times that of PoE.
	
	\begin{defn}
		\textit{When $n \to \infty$, let $\bm{X} \in R^{n \times d}$ be dense in $\Omega \in [0,1]^d$ such that for any $\bm{x} \in \Omega$ we have $\lim_{n \to \infty} \min_{1\le i\le n} \|\bm{x}_i - \bm{x} \| = 0$. Besides, the underlying function to be approximated has true continuous response $\mu_{\eta}(\bm{x})$ and true noise variance $\sigma^2_{\eta}$.}
	\end{defn}
	
	Firstly, for the disjoint partition that uses clustering techniques to partition the data $\mathcal{D}$ into disjoint local subsets $\{\mathcal{D}_i\}_{i=1}^M$, The proposition below reveals that when $n \to \infty$, PoE and (R)BCM produce overconfident prediction variance that shrinks to zero; on the contrary, GPoE provides conservative prediction variance.
	
	\begin{prop} \label{Prop_Typical_disjoint}
		Let $\{\mathcal{D}_i\}_{i=1}^{M_n}$ be a disjoint partition of the training data $\mathcal{D}$. Let the expert $\mathcal{M}_i$ trained on $\mathcal{D}_i$ be GP with zero mean and stationary covariance function $k(.) > 0$. We further assume that (i) $\lim_{n \to \infty} M_n = \infty$ and (ii) $\lim_{n \to \infty} n/M_n^2 > 0$, where the second condition implies that the subset size $m_0 = n/M_n$ and the number of experts $M_n$ are comparable such that too weak experts are not preferred. Besides, from the second condition we have $m_0 \to_{n \to \infty} \infty$, which implies that the experts become more informative with increasing $n$. Then, PoE and (R)BCM produce overconfident prediction variance at $\bm{x}_* \in \Omega$ as
		\begin{equation}
		\lim_{n \to \infty} \sigma^2_{\mathcal{A},n}(\bm{x}_*) = 0,
		\end{equation}
		whereas GPoE yields conservative prediction variance
		\begin{equation}
		\sigma^{2}_{\eta} < \lim_{n \to \infty} \sigma^2_{\mathcal{A},n}(\bm{x}_*) < \sigma^2_{b_n}(\bm{x}_*) < \sigma^2_{**},
		\end{equation} 
		where $\sigma^2_{b_n}(\bm{x}_*)$ is offered by the farthest expert $\mathcal{M}_{b_n}$ ($1 \le b_n \le M_n$) whose prediction variance is closet to $\sigma^2_{**}$.
	\end{prop}
	The detailed proof is given in Appendix~\ref{App_Prop_Typical_disjoint}. Moreover, we have the following findings.
	
	\begin{rem} \label{Rem_1}
		\textit{For the averaging $\sigma^{-2}_{\mathrm{GPoE}} = \frac{1}{M} \sum_{i=1}^M \sigma^{-2}_i$ and $\mu_{\mathrm{(G)PoE}} = \sum_{i=1}^M \frac{\sigma^{-2}_i}{\sum \sigma^{-2}_i} \mu_i$ using disjoint partition, more and more experts become relatively far away from $\bm{x}_*$ when $n \to \infty$, i.e., the prediction variances at $\bm{x}_*$ approach $\sigma^2_{**}$ and the prediction means approach the prior mean $\mu_{**}$. Hence, empirically, when $n \to \infty$, the conservative $\sigma^{2}_{\mathrm{GPoE}}$ approaches $\sigma^2_{b_n}$, and the $\mu_{\mathrm{(G)PoE}}$ approaches $\mu_{**}$.}
	\end{rem}
	
	\begin{rem} \label{Rem_2}
		\textit{The BCM's prediction variance is always larger than that of PoE since 
			\begin{equation*}
			a_* = \frac{\sigma^{-2}_{\mathrm{PoE}}(\bm{x}_*)}{\sigma^{-2}_{\mathrm{BCM}}(\bm{x}_*)} =  \frac{\sum_{i=1}^M \sigma^{-2}_{i}(\bm{x}_*)}{\sum_{i=1}^M \sigma^{-2}_{i}(\bm{x}_*) - (M-1)\sigma^{-2}_{**}} > 1
			\end{equation*}
			for $M > 1$. This means $\sigma^{2}_{\mathrm{PoE}}$ deteriorates faster to zero when $n \to \infty$. Besides, it is observed that $\mu_{\mathrm{BCM}}$ is $a_*$ times that of PoE, which alleviates the deterioration of prediction mean when $n \to \infty$. However, when $\bm{x}_*$ is leaving $\bm{X}$, $a_* \to M$ since $\sigma^{-2}_i(\bm{x}_*) \to \sigma^{-2}_{**}$. That is why BCM suffers from undesirable prediction mean when leaving $\bm{X}$.}
	\end{rem}
	
	Secondly, for the random partition that assigns the data points randomly to the experts without replacement, The proposition below implies that when $n \to \infty$, the prediction variances of PoE and (R)BCM will shrink to zero; the PoE's prediction mean will recover $\mu_{\eta}(\bm{x})$, but the (R)BCM's prediction mean cannot; interestingly, the simple GPoE can converge to the underlying true predictive distribution.
	
	\begin{prop} \label{Prop_Typical_random}		
		Let $\{\mathcal{D}_i\}_{i=1}^{M_n}$ be a random partition of the training data $\mathcal{D}$ with (i) $\lim_{n \to \infty} M_n = \infty$ and (ii) $\lim_{n \to \infty} n/M_n^2 > 0$. Let the experts $\{\mathcal{M}_i\}_{i=1}^{M_n}$ be GPs with zero mean and stationary covariance function $k(.) > 0$. Then, for the aggregated predictions at $\bm{x}_* \in \Omega$ we have
		\begin{equation}
		\left\lbrace \begin{aligned}
		&\lim_{n \to \infty} \mu_{\mathrm{PoE}}(\bm{x}_*) = \mu_{\eta}(\bm{x}_*), \, \lim_{n \to \infty} \sigma^2_{\mathrm{PoE}}(\bm{x}_*) = 0, \\
		&\lim_{n \to \infty} \mu_{\mathrm{GPoE}}(\bm{x}_*) = \mu_{\eta}(\bm{x}_*), \, \lim_{n \to \infty} \sigma^2_{\mathrm{GPoE}}(\bm{x}_*) = \sigma^2_{\eta}, \\
		&\lim_{n \to \infty} \mu_{\mathrm{(R)BCM}}(\bm{x}_*) = a \mu_{\eta}(\bm{x}_*), \, \lim_{n \to \infty} \sigma^2_{\mathrm{(R)BCM}}(\bm{x}_*) = 0, \\
		\end{aligned}
		\right.
		\end{equation}
		
		where $a = \sigma_{\eta}^{-2}/(\sigma_{\eta}^{-2} - \sigma_{**}^{-2}) \geq 1$ and the equality holds when $\sigma^2_{\eta} = 0$.
	\end{prop}
	
	The detailed proof is provided in Appendix~\ref{App_Prop_Typical_random}. Propositions~\ref{Prop_Typical_disjoint} and~\ref{Prop_Typical_random} imply that no matter what kind of data partition has been used, the prediction variances of PoE and (R)BCM will shrink to zero when $n \to \infty$, which strictly limits their usability since no benefits can be gained from such useless uncertainty information.
	
	As for data partition, intuitively, the random partition provides overlapping and coarse global information about the target function, which limits the ability to describe quick-varying characteristics. On the contrary, the disjoint partition provides separate and refined local information, which enables the model to capture the variability of target function. The superiority of disjoint partition has been empirically confirmed in \cite{rulliere2017nested}. Therefore, unless otherwise indicated, we employ disjoint partition for the aggregation models throughout the article.
	
	As for time complexity, the five aggregation models have the same training process, and they only differ in how to combine the experts' predictions. For (G)PoE and (R)BCM, their time complexity in prediction scales as $\mathcal{O}(nm_0^2) + \mathcal{O}(n'nm_0)$ where $n'$ is the number of test points.\footnote{$\mathcal{O}(nm_0^2)$ is induced by the update of $M$ GP experts after optimizing hyperparameters.} For the complicated NPAE, it however needs to invert an $M \times M$ matrix $\bm{K}_{\mathcal{A}}$ at each test point, leading to a greatly increased time complexity in prediction as $\mathcal{O}(n'n^2)$.\footnote{The predicting complexity of NPAE can be reduced by employing various hierarchical computing structure \cite{rulliere2017nested}, which however cannot provide identical predictions.}
	
	The inconsistency of (G)PoE and (R)BCM and the extremely time-consuming process of NPAE impose the demand of developing a consistent yet efficient aggregation model for large-scale GP regression.

	\section{Generalized robust Bayesian committee machine} \label{Sec_3}
	\subsection{GRBCM}
	Our proposed GRBCM divides $M$ experts into two groups. The first group has a \textit{global communication expert} $\mathcal{M}_c$ trained on the subset $\mathcal{D}_c = \mathcal{D}_1$, and the second group contains the remaining $M-1$ global or local experts\footnote{``Global'' means the expert is trained on a random subset, whereas ``local'' means it is trained on a disjoint subset.} $\{\mathcal{M}_i\}_{i=2}^M$ trained on $\{\mathcal{D}_i\}_{i=2}^M$, respectively. The training process of GRBCM is identical to that of typical aggregations in section~\ref{Sec_2.1}. The prediction process of GRBCM, however, is different. Particularly, GRBCM assigns the global communication expert with the following properties:
	\begin{itemize}
		\item (\textit{Random selection}) The communication subset $\mathcal{D}_c$ is a random subset wherein the points are randomly selected without replacement from $\mathcal{D}$. It indicates that the points in $\bm{X}_c$ spread over the entire domain, which enables $\mathcal{M}_c$ to capture the main features of the target function. Note that there is no limit to the partition type for the remaining $M-1$ subsets.
		\item (\textit{Expert communication}) The expert $\mathcal{M}_c$ with predictive distribution $p_c(y_*|\mathcal{D}_c,\bm{x}_*)\sim \mathcal{N}(\mu_c,\sigma^2_c)$ is allowed to communicate with each of the remaining experts $\{\mathcal{M}_i\}_{i=2}^M$. It means we can utilize the augmented data $\mathcal{D}_{+i} = \{\mathcal{D}_c, \mathcal{D}_i\}$ to improve over the base expert $\mathcal{M}_c$, leading to a new expert $\mathcal{M}_{+i}$ with the improved predictive distribution as $p_{+i}(y_*|\mathcal{D}_{+i},\bm{x}_*)\sim \mathcal{N}(\mu_{+i},\sigma^2_{+i})$ for $2 \le i \le M$.
		\item (\textit{Conditional independence}) Given the communication subset $\mathcal{D}_c$ and $y_*$, the independence assumption $\mathcal{D}_i \perp \mathcal{D}_j|\mathcal{D}_c,y_*$ holds for $2 \le i \ne j \le M$.
	\end{itemize} 
	
	Given the conditional independence assumption and the weights $\{\beta_i\}_{i=2}^M$, we approximate the exact predictive distribution $p(y_*|\mathcal{D}, \bm{x}_*)$ using the Bayes rule as
	\begin{equation} \label{Eq_Bayes_GRBCM}
	\begin{aligned}
	&p(y_*|\mathcal{D}, \bm{x}_*) \\
	&\propto p(y_*|\bm{x}_*)p(\mathcal{D}_c|y_*,\bm{x}_*)  \prod_{i=2}^M p(\mathcal{D}_i|\{\mathcal{D}_j\}^{i-1}_{j=1},y_*,\bm{x}_*) \\
	&\approx p(y_*|\bm{x}_*)p(\mathcal{D}_c|y_*,\bm{x}_*) \prod_{i=2}^{M} p^{\beta_i}(\mathcal{D}_i|\mathcal{D}_c,y_*,\bm{x}_*) \\
	&= \frac{p(y_*|\bm{x}_*) \prod_{i=2}^{M} p^{\beta_i}(\mathcal{D}_{+i}|y_*,\bm{x}_*)}{p^{\sum_{i=2}^M \beta_i -1}(\mathcal{D}_c|y_*,\bm{x}_*)}. \\
	\end{aligned}
	\end{equation}
	Note that $p(\mathcal{D}_2|\mathcal{D}_c,y_*,\bm{x}_*)$ is exact with no approximation in \eqref{Eq_Bayes_GRBCM}. Hence, we set $\beta_2 = 1$.
	
\begin{figure}[ht]
		\vskip 0.0in
		\begin{center}
			\centerline{\includegraphics[width=1.0\columnwidth]{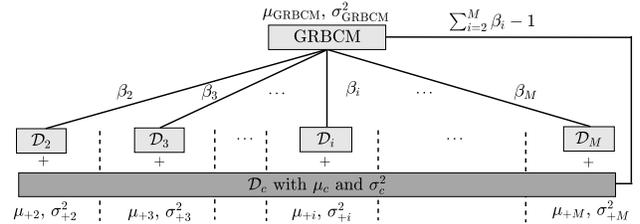}}
			\caption{The GRBCM aggregation model.}
			\label{Fig_GRBCM}
		\end{center}
		\vskip -0.3in
	\end{figure}

	With \eqref{Eq_Bayes_GRBCM}, GRBCM's predictive distribution is
	\begin{equation} \label{Eq_GRBCM}
	\begin{aligned}
	p_{\mathcal{A}}(y_*|\mathcal{D},\bm{x}_*) &= \frac{\prod_{i=2}^{M} p_{+i}^{\beta_i}(y_*|\mathcal{D}_{+i}, \bm{x}_*)}{p_c^{\sum_{i=2}^M \beta_i -1}(y_*|\mathcal{D}_c,\bm{x}_*)}.
	\end{aligned}
	\end{equation}
	with
	\begin{subequations}
		\begin{align}
		\mu_{\mathcal{A}}(\bm{x}_*) &= \sigma_{\mathcal{A}}^2(\bm{x}_*) \left[ \sum_{i=2}^M \beta_i \sigma_{+i}^{-2}(\bm{x}_*) \mu_{+i}(\bm{x}_*) \right. \nonumber \\
		&\left. - \left(\sum_{i=2}^M \beta_i -1\right)\sigma_c^{-2}(\bm{x}_*)\mu_c(\bm{x}_*) \right], \label{Eq_mu_GRBCM} \\
		\sigma_{\mathcal{A}}^{-2}(\bm{x}_*) &= \sum_{i=2}^M \beta_i \sigma_{+i}^{-2}(\bm{x}_*)  - \left(\sum_{i=2}^M \beta_i -1\right)\sigma_c^{-2}(\bm{x}_*). \label{Eq_s2_GRBCM} 
		\end{align}
	\end{subequations}
	Different from (R)BCM, GRBCM employs the informative $\sigma^{-2}_c$ rather than the prior $\sigma^{-2}_{**}$  to correct the prediction precision in \eqref{Eq_s2_GRBCM}, leading to consistent predictions when $n \to \infty$, which will be proved below. Also, the prediction mean of GRBCM in~\eqref{Eq_mu_GRBCM} now is corrected by $\mu_c(\bm{x}_*)$. Fig.~\ref{Fig_GRBCM} depicts the structure of the GRBCM aggregation model.
	
	In~\eqref{Eq_mu_GRBCM} and~\eqref{Eq_s2_GRBCM}, the parameter $\beta_i$ ($i > 2$) akin to that of RBCM is defined as the difference in the differential entropy between the base predictive distribution $p_c(y_*|\mathcal{D}_c,\bm{x}_*)$ and  the enhanced predictive distribution $p_{+i}(y_*|\mathcal{D}_{+i},\bm{x}_*)$ as
	\begin{equation} \label{Eq_GRBCM_beta}
	\beta_i = \left\lbrace 
	\begin{aligned}
	&1, &i = 2, \\
	&0.5(\log\sigma_{c}^2(\bm{x}_*) - \log\sigma_{+i}^2(\bm{x}_*)), &3 \le i \le M.
	\end{aligned}
	\right.
	\end{equation}
	It is found that after adding a subset $\mathcal{D}_i$ ($i \geq 2$) into the communication subset $\mathcal{D}_c$, if there is little improvement of $p_{+i}(y_*|\mathcal{D}_{+i},\bm{x}_*)$ over $p_c(y_*|\mathcal{D}_c,\bm{x}_*)$, we weak the vote of $\mathcal{M}_{+i}$ by assigning a small $\beta_i$ that approaches zero. 
	
	As for the size of $\bm{X}_c$, more data points bring more informative $\mathcal{M}_c$ and better GRBCM predictions at the cost of higher computing complexity. In this article, we assign all the experts with the same training size as $n_c = n_i = m_0$ and $n_{+i} = 2m_0$ for $2 \le i \le M$.
	
	Next, we show that the GRBCM's predictive distribution will converge to the underlying true predictive distribution when $n \to \infty$.
	\begin{prop} \label{Prop_GRBCM}		
		Let $\{\mathcal{D}_i\}_{i=1}^{M_n}$ be a partition of the training data $\mathcal{D}$ with (i) $\lim_{n \to \infty} M_n = \infty$ and (ii) $\lim_{n \to \infty} n/M_n^2 > 0$. Besides, among the $M$ subsets, there is a global communication subset $\mathcal{D}_c$, the points in which are randomly selected from $\mathcal{D}$ without replacement. Let the global expert $\mathcal{M}_c$ and the enhanced experts $\{\mathcal{M}_{+i}\}_{i=2}^{M_n}$ be GPs with zero mean and stationary covariance function $k(.)>0$. Then, GRBCM yields consistent predictions as
		\begin{equation}
		\left\lbrace \begin{aligned}
		\lim_{n \to \infty} \mu_{\mathrm{GRBCM}}(\bm{x}_*) &= \mu_{\eta}(\bm{x}_*), \\
		\lim_{n \to \infty}  \sigma^2_{\mathrm{GRBCM}}(\bm{x}_*) &= \sigma^2_{\eta}.
		\end{aligned}
		\right.
		\end{equation}
	\end{prop}
	
	The detailed proof is provided in Appendix~\ref{App_Prop_GRBCM}. It is found in Proposition~\ref{Prop_GRBCM} that apart from the requirement that the communication subset $\mathcal{D}_c$ should be a random subset, the consistency of GRBCM holds for any partition of the remaining data $\mathcal{D} \backslash \mathcal{D}_c$. Besides, according to Propositions~\ref{Prop_Typical_random} and~\ref{Prop_GRBCM}, both GPoE and GRBCM produce consistent predictions using random partition. It is known that the GP model $\mathcal{M}$ provides more confident predictions, i.e., lower uncertainty $U(\mathcal{M}) = \int \sigma^2({\bm{x}}) d \bm{x}$, with more data points. Since GRBCM trains experts on more informative subsets $\{\mathcal{D}_{+i}\}_{i=2}^M$, we have the following finding.
	\begin{rem} \label{Rem_3}
		\textit{When using random subsets, the GRBCM's prediction uncertainty is always lower than that of GPoE, since the discrepancy $\delta_{U^{-1}} = U_{\mathrm{GRBCM}}^{-1} - U_{\mathrm{GPoE}}^{-1}$ satisfies
			\begin{equation*}
			\begin{aligned}			
			\delta_{U^{-1}} =& \left[U^{-1}(\mathcal{M}_{+2}) - \frac{1}{M_n} \sum_{i=1}^{M_n} U^{-1}(\mathcal{M}_{i})\right] \\
			+& \int \sum_{i=3}^{M_n} \beta_i \left(\sigma_{+i}^{-2}(\bm{x}_*)  - \sigma_c^{-2}(\bm{x}_*) \right) d\bm{x}_* > 0
			\end{aligned}
			\end{equation*}
			for a large enough $n$. It means compared to GPoE, GRBCM converges faster to the underlying function when $n \to \infty$.}
	\end{rem}
	
	Finally, similar to RBCM, GRBCM can be executed in multi-layer computing architectures with identical predictions \cite{deisenroth2015distributed, ionescu2015revisiting}, which allow to run optimally and efficiently with the available computing infrastructure for distributed computing.
	
	\subsection{Complexity}
	Assuming that the experts $\{\mathcal{M}_i\}_{i=1}^M$ have the same training size $n_i = m_0 =  n/M$ for $1 \le i \le M$. Compared to (G)PoE and (R)BCM, the proposed GRBCM has a higher time complexity in prediction due to the construction of new experts $\{\mathcal{M}_{+i}\}_{i=2}^M$. In prediction, it first needs to calculate the inverse of $k(\bm{X}_c, \bm{X}_c)$ and $M-1$ augmented covariance matrices $\{k(\{\bm{X}_i,\bm{X}_c\},\{\bm{X}_i,\bm{X}_c\}) \}_{i=2}^M$, which scales as $\mathcal{O}(8nm_0^2 - 7m_0^3)$, in order to obtain the predictions $\mu_c$, $\{\mu_{+i}\}_{i=2}^M$ and $\sigma^2_c$, $\{\sigma^2_{+i}\}_{i=2}^M$. Then, it combines the predictions of $\mathcal{M}_c$ and $\{\mathcal{M}_{+i}\}_{i=2}^M$ at $n'$ test points. Therefore, the time complexity of the GRBCM prediction process is $\mathcal{O}(\alpha nm_0^2) + \mathcal{{O}}(\beta n'nm_0)$, where $\alpha=(8M-7)/M$ and $\beta=(4M-3)/M$.

	\section{Numerical experiments} \label{Sec_4}
	\subsection{Toy example}
	We employ a 1D toy example
	\begin{equation} \label{Eq_toy}
	\begin{aligned}
	f(x) &= 5x^2 \sin(12x) + (x^3-0.5) \sin(3x-0.5) \\
	& + 4 \cos(2x) + \epsilon ,
	\end{aligned}
	\end{equation}
	where $\epsilon \sim \mathcal{N}(0,0.25)$, to illustrate the characteristics of existing aggregation models.
	
	We generate $n=10^4$, $5 \times 10^4$, $10^5$, $5 \times 10^5$ and $10^6$ training points, respectively, in $[0,1]$, and select $n' = 0.1n$ test points randomly in $[-0.2,1.2]$. We pre-normalize each column of $\bm{X}$ and $\bm{y}$ to zero mean and unit variance. Due to the global expert $\mathcal{M}_c$ in GRBCM, we slightly modify the disjoint partition: we first generate a random subset and then use the \textit{k}-means technique to generate $M-1$ disjoint subsets. Each expert is assigned with $m_0 = 500$ data points. We implement the aggregations by the GPML toolbox\footnote{\url{http://www.gaussianprocess.org/gpml/code/matlab/doc/}} using the SE kernel in~\eqref{Eq_SE_Kernel} and the conjugate gradients algorithm with the maximum number of evaluations as 500, and execute the code on a workstation with four 3.70 GHz cores and 16 GB RAM (multi-core computing in Matalb is employed). Finally, we use the Standardized Mean Square Error (SMSE) to evaluate the accuracy of prediction mean, and the Mean Standardized Log Loss (MSLL) to quantify the quality of predictive distribution \cite{rasmussen2006gaussian}.
	
	\begin{figure}[ht]
		\vskip 0.0in
		\begin{center}
			\centerline{\includegraphics[width=0.9\columnwidth]{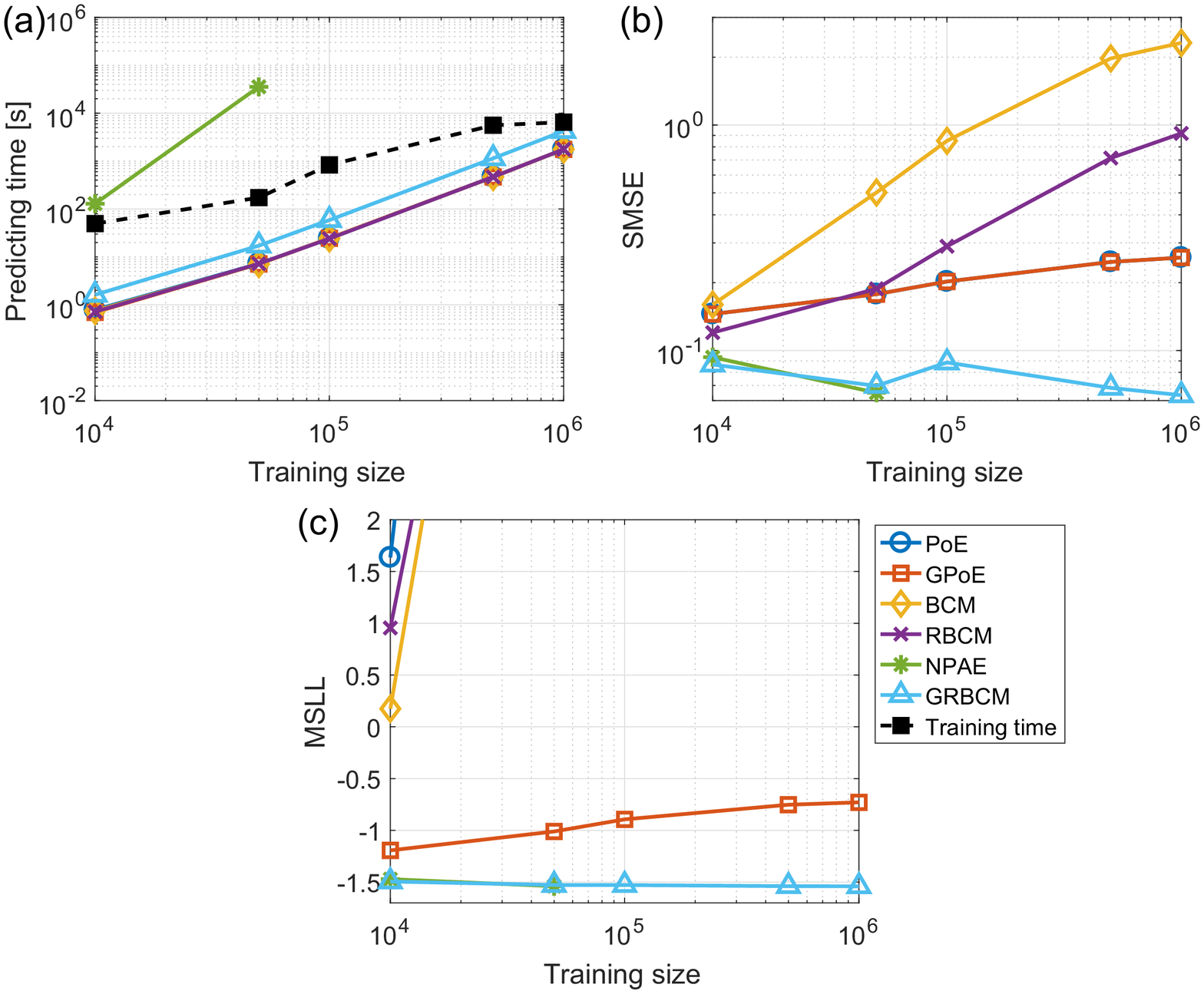}}
			\caption{Comparison of different aggregation models on the toy example in terms of (a) computing time, (b) SMSE and (c) MSLL.}
			\label{Fig_ResultsAnalysis_Toy}
		\end{center}
		\vskip -0.3in
	\end{figure}
	
	Fig.~\ref{Fig_ResultsAnalysis_Toy} depicts the comparative results of six aggregation models on the toy example. Note that NPAE using $n > 5 \times 10^4$ is unavailable due to the time-consuming prediction process. Fig.~\ref{Fig_ResultsAnalysis_Toy}(a) shows that these models require the same training time, but they differ in the predicting time. Due to the communication expert, the GRBCM's predicting time slightly offsets the curves of (G)PoE and (R)BCM. The NPAE however exhibits significantly larger predicting time with increasing $M$ and $n'$. Besides, Fig.~\ref{Fig_ResultsAnalysis_Toy}(b) and (c) reveal that GRBCM and NPAE yield better predictions with increasing $n$, which confirm their consistency when $n \to \infty$.\footnote{Further discussions of GRBCM is shown in Appendix~\ref{App_GRBCM_toy}.} As for NPAE, though performing slightly better than GRBCM using $n = 5 \times 10^4$, it requires several orders of magnitude larger predicting time, rendering it unsuitable for cases with many test points and subsets.
	
	\begin{figure}[ht]
		\vskip -0.05in
		\begin{center}
			\centerline{\includegraphics[width=0.9\columnwidth]{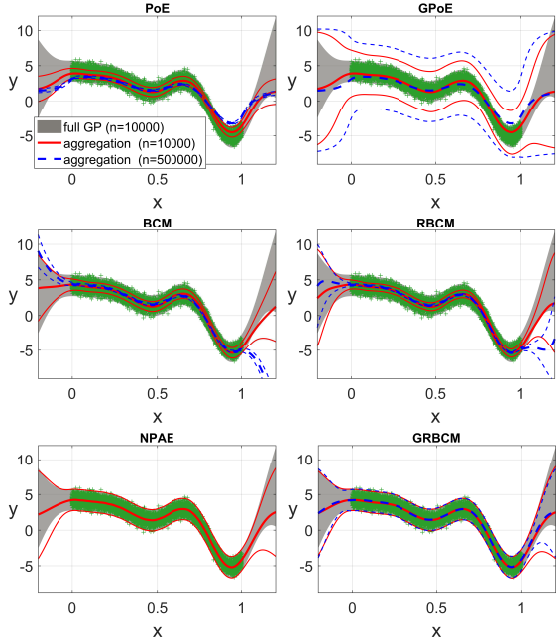}}
			\caption{Illustrations of the aggregation models on the toy example. The green ``+'' symbols represent the $10^4$ data points. The shaded area indicates 99\% confidence intervals of the full GP predictions using $n=10^4$.}
			\label{Fig_Toy_comparison}
		\end{center}
		\vskip -0.3in
	\end{figure}
	
	Fig.~\ref{Fig_Toy_comparison} illustrates the six aggregation models using $n = 10^4$ and $n = 5 \times 10^5$, respectively, in comparison to the full GP (ground truth) using $n=10^4$.\footnote{The full GP is intractable using our computer for $n = 5 \times 10^5$.} It is observed that in terms of prediction mean, as discussed in remark~\ref{Rem_1}, PoE and GPoE provide poorer results in the entire domain with increasing $n$. On the contrary, BCM and RBCM provide good predictions in the range $[0,1]$. As discussed in remark~\ref{Rem_2}, BCM however yields unreliable predictions when leaving the training data. RBCM alleviates the issue by using a varying $\beta_i$. In terms of prediction variance, with increasing $n$, PoE and (R)BCM tend to shrink to zero (overconfident), while GPoE tends to approach $\sigma^2_{**}$ (too conservative). Particularly, PoE always has the largest MSLL value in Fig.~\ref{Fig_ResultsAnalysis_Toy}(b), since as discussed in remark~\ref{Rem_2}, its prediction variance approaches zero faster.
	
	\subsection{Medium-scale datasets} \label{Sec_4.2}
	We use two realistic datasets, \textit{kin40k} (8D, $10^4$ training points, $3 \times 10^4$ test points) \cite{seeger2003fast} and \textit{sarcos} (21D, 44484 training points, 4449 test points) \cite{rasmussen2006gaussian}, to assess the performance of our approach.
	
	The comparison includes all the aggregations except the expensive NPAE.\footnote{The comparison of NPAE and GRBCM are separately provided in Appendix~\ref{App_results_NPAE}.} Besides, we employ the fully independent training conditional (FITC) \cite{snelson2006sparse}, the GP using stochastic variational inference (SVI)\footnote{\url{https://github.com/SheffieldML/GPy}} \cite{hensman2013gaussian}, and the subset-of-data (SOD) \cite{chalupka2013framework} for comparison. We select the inducing size $m$ for FITC and SVI, the batch size $m_b$ for SVI, and the subset size $m_{\mathrm{sod}}$ for SOD, such that the computing time is similar to or a bit larger than that of GRBCM. Particularly, we choose $m = 200$, $m_b = 0.1n$ and $m_{\mathrm{sod}} = 2500$ for \textit{kin40k}, and $m = 300$, $m_b = 0.1n$ and $m_{\mathrm{sod}} = 3000$ for \textit{sarcos}. Differently, SVI employs the stochastic gradients algorithm with $t_{\mathrm{sg}}=1200$ iterations. Finally, we adopt the disjoint partition used before to divide the \textit{kin40k} dataset into 16 subsets, and the \textit{sarcos} dataset into 72 subsets for the aggregations. Each experiment is repeated ten times.
	
	\begin{figure}[ht]
		\vskip -0.0in
		\begin{center}
			\centerline{\includegraphics[width=0.9\columnwidth]{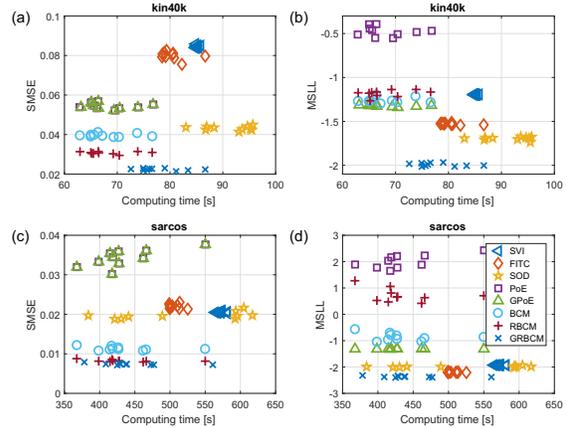}}
			\caption{Comparison of the approximation models on the \textit{kin40k} and \textit{sarcos} datasets.}
			\label{Fig_Comparison_kin40k_sarcos}
		\end{center}
		\vskip -0.3in
	\end{figure}
	
	Fig.~\ref{Fig_Comparison_kin40k_sarcos} depicts the comparative results of different approximation models over 10 runs on the \textit{kin40k} and \textit{sarcos} datasets. The horizontal axis represents the sum of training and predicting time. It is first observed that GRBCM provides the best performance on the two datasets in terms of both SMSE and MSLL at the cost of requiring a bit more computing time than (G)PoE and (R)BCM. As for (R)BCM, the small SMSE values reveal that they provide better prediction mean than FITC and SOD; but the large MSLL values again confirm that they provide overconfident prediction variance. As for (G)PoE, they suffer from poor prediction mean, as indicated by the large SMSE; but GPoE performs well in terms of MSLL. Finally, the simple SOD outperforms FITC and SVI on the \textit{kin40k} dataset, and performs similarly on the \textit{sarcos} dataset, which are consistent with the findings in \cite{chalupka2013framework}.
	
	\begin{figure}[ht]
		\vskip -0.0in
		\begin{center}
			\centerline{\includegraphics[width=0.9\columnwidth]{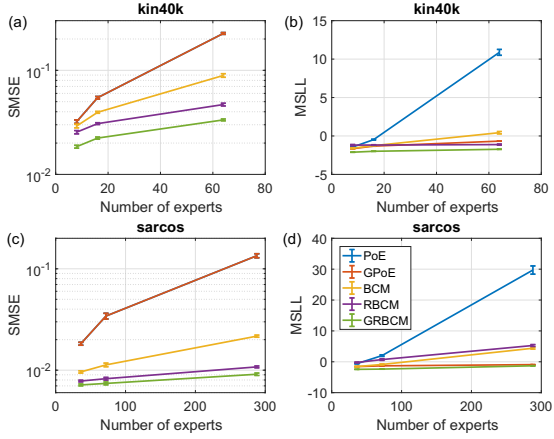}}
			\caption{Comparison of the aggregation models using different numbers of experts on the \textit{kin40k} and \textit{sarcos} datasets.}
			\label{Fig_Comparison_kin40k_sarcos_M}
		\end{center}
		\vskip -0.3in
	\end{figure}
	
	Next, we explore the impact of the number $M$ of experts on the performance of aggregations.  To this end, we run them on the \textit{kin40k} dataset with $M$ respectively being 8, 16 and 64, and we run on the \textit{sarcos} dataset with $M$ respectively being 36, 72 and 288. The results in Fig.~\ref{Fig_Comparison_kin40k_sarcos_M} turn out that all the aggregations perform worse with increasing $M$, since the experts become weaker; but GRBCM still yields the best performance with different $M$. Besides, with increasing $M$, the poor prediction mean and the vanishing prediction variance of PoE result in the sharp increase of MSLL values.
	
	\begin{figure}[ht]
		\vskip -0.0in
		\begin{center}
			\centerline{\includegraphics[width=0.9\columnwidth]{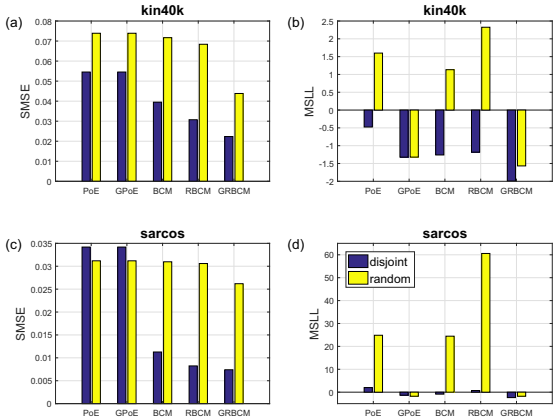}}
			\caption{Comparison of the aggregation models using disjoint and random partitions on the \textit{kin40k} dataset ($M = 16$) and the \textit{sarcos} dataset ($M = 72$).}
			\label{Fig_Comparison_kin40k_sarcos_partition}
		\end{center}
		\vskip -0.3in
	\end{figure}
	
	Finally, we investigate the impact of data partition (disjoint or random) on the performance of aggregations. The average results in Fig.~\ref{Fig_Comparison_kin40k_sarcos_partition} turn out that the disjoint partition is more beneficial for the aggregations. The results are expectable since the disjoint subsets provide separate and refined local information, whereas the random subsets provide overlapping and coarse global information. But we observe that GPoE performs well on the \textit{sarcos} dataset using random partition, which confirms the conclusions in Proposition~\ref{Prop_Typical_random}. Besides, as revealed in remark~\ref{Rem_3}, even using random partition, GRBCM outperforms GPoE.

	\subsection{Large-scale datasets} \label{Sec_4.3}
	This section explores the performance of aggregations and SVI on two large-scale datasets. We first assess them on the 90D \textit{song} dataset, which is a subset of the million song dataset \cite{bertin2011million}. The \textit{song} dataset is partitioned into 450000 training points and 65345 test points. We then assess the models on the 11D \textit{electric} dataset that is partitioned into 1.8 million training points and 249280 test points. We follow the normalization and data pre-processing in \cite{wilson2016deep} to generate the two datasets.\footnote{The datasets and the pre-processing scripts are available in \url{https://people.orie.cornell.edu/andrew/}.} For the \textit{song} dataset, we use the foregoing disjoint partition to divide it into $M=720$ subsets, and use $m = 800$, $m_b = 5000$ and $t_{\mathrm{sg}}=1300$ for SVI; for the \textit{electric} dataset, we divide it into $M=2880$ subsets, and use $m = 1000$, $m_b = 5000$ and $t_{\mathrm{sg}}=1500$ for SVI. As a result, each expert is assigned with $m_0 = 625$ data points for the aggregations.
	
	\begin{table}[t]
		\caption{Comparative results of the aggregation models and SVI on the \textit{song} and \textit{electric} datasets.}
		\label{Tab_LargeScaleResults}
		\vskip -0.1in
		\begin{center}
			\begin{small}
				\begin{sc}
					\begin{tabular}{lcccc}
						\toprule
						& \multicolumn{2}{c}{\textit{song} (450K)} & \multicolumn{2}{c}{\textit{electric} (1.8M)} \\
						\hline
						& SMSE & MSLL & SMSE & MSLL \\
						\midrule
						PoE        & 0.8527 & 328.82 & 0.1632 & 1040.3 \\
						GPoE     & 0.8527 & 0.1159 & 0.1632 & 24.940 \\
						BCM      & 2.6919 & 156.62 & 0.0073 & 51.081 \\
						RBCM    & 1.3383 & 24.930 & 0.0027 & 85.657 \\
						SVI & 0.7909 & \textbf{-0.1885} & 0.0042 & -1.1410 \\
						GRBCM & \textbf{0.7321} &-0.1571 & \textbf{0.0024} & \textbf{-1.3161} \\
						\bottomrule
					\end{tabular}
				\end{sc}
			\end{small}
		\end{center}
		\vskip -0.3in
	\end{table}
	
	Table~\ref{Tab_LargeScaleResults} reveals that the (G)PoE's SMSE value is smaller than that of (R)BCM on the \textit{song} dataset. The poor prediction mean of BCM is caused by the fact that the \textit{song} dataset is highly clustered such that BCM suffers from weak experts in regions with scarce points. On the contrary, due to the almost uniform distribution of the \textit{electric} data points, the (R)BCM's SMSE is much smaller than that of (G)PoE. Besides, unlike the vanishing prediction variances of PoE and (R)BCM when $n \to \infty$, GPoE provides conservative prediction variance, resulting in small MSLL values on the two datasets. The proposed GRBCM always outperforms the other aggregations in terms of both SMSE and MSLL on the two datasets due to the consistency. Finally, GRBCM performs similarly to SVI on the \textit{song} dataset; but GRBCM outperforms SVI on the \textit{electric} dataset.

	\section{Conclusions} \label{Sec_5}
	To scale the standard GP to large-scale regression, we present the GRBCM aggregation model, which introduces a global communication expert to yield consistent yet efficient predictions when $n \to \infty$. Through theoretical and empirical analyses, we demonstrated the superiority of GRBCM over existing aggregations on datasets with up to 1.8M training points.
	
	The superiority of local experts is the capability of capturing local patterns. Hence, further works will consider the experts with individual hyperparameters in order to capture non-stationary and heteroscedastic features.
	
	% Acknowledgements should only appear in the accepted version.
	\section*{Acknowledgements}
	This work was conducted within the Rolls-Royce@NTU Corporate Lab with support from the National Research Foundation (NRF) Singapore under the Corp Lab@University Scheme. It is also partially supported by the Data Science and Artificial Intelligence Research Center (DSAIR) and the School of Computer Science and Engineering at Nanyang Technological University.

	% In the unusual situation where you want a paper to appear in the
	% references without citing it in the main text, use \nocite
	
	\bibliography{GRBCM_ICML}

\begin{thebibliography}{37}
\providecommand{\natexlab}[1]{#1}
\providecommand{\url}[1]{\texttt{#1}}
\expandafter\ifx\csname urlstyle\endcsname\relax
  \providecommand{\doi}[1]{doi: #1}\else
  \providecommand{\doi}{doi: \begingroup \urlstyle{rm}\Url}\fi

\bibitem[Alvarez et~al.(2012)Alvarez, Rosasco, Lawrence,
  et~al.]{alvarez2012kernels}
Alvarez, Mauricio~A, Rosasco, Lorenzo, Lawrence, Neil~D, et~al.
\newblock Kernels for vector-valued functions: {A} review.
\newblock \emph{Foundations and Trends{\textregistered} in Machine Learning},
  4\penalty0 (3):\penalty0 195--266, 2012.

\bibitem[Bauer et~al.(2016)Bauer, van~der Wilk, and
  Rasmussen]{bauer2016understanding}
Bauer, Matthias, van~der Wilk, Mark, and Rasmussen, Carl~Edward.
\newblock Understanding probabilistic sparse {G}aussian process approximations.
\newblock In \emph{Advances in Neural Information Processing Systems}, pp.\
  1533--1541. Curran Associates, Inc., 2016.

\bibitem[Bertin-Mahieux et~al.(2011)Bertin-Mahieux, Ellis, Whitman, and
  Lamere]{bertin2011million}
Bertin-Mahieux, Thierry, Ellis, Daniel~PW, Whitman, Brian, and Lamere, Paul.
\newblock The million song dataset.
\newblock In \emph{ISMIR}, pp.\  1--6, 2011.

\bibitem[Bui \& Turner(2014)Bui and Turner]{bui2014tree}
Bui, Thang~D and Turner, Richard~E.
\newblock Tree-structured {G}aussian process approximations.
\newblock In \emph{Advances in Neural Information Processing Systems}, pp.\
  2213--2221. Curran Associates, Inc., 2014.

\bibitem[Cao \& Fleet(2014)Cao and Fleet]{cao2014generalized}
Cao, Yanshuai and Fleet, David~J.
\newblock Generalized product of experts for automatic and principled fusion of
  {G}aussian process predictions.
\newblock \emph{arXiv preprint arXiv:1410.7827}, 2014.

\bibitem[Chalupka et~al.(2013)Chalupka, Williams, and
  Murray]{chalupka2013framework}
Chalupka, Krzysztof, Williams, Christopher~KI, and Murray, Iain.
\newblock A framework for evaluating approximation methods for {G}aussian
  process regression.
\newblock \emph{Journal of Machine Learning Research}, 14\penalty0
  (Feb):\penalty0 333--350, 2013.

\bibitem[Choi \& Schervish(2004)Choi and Schervish]{choi2004posterior}
Choi, Taeryon and Schervish, Mark~J.
\newblock Posterior consistency in nonparametric regression problems under
  {G}aussian process priors.
\newblock Technical report, Carnegie Mellon University, 2004.

\bibitem[Deisenroth \& Ng(2015)Deisenroth and Ng]{deisenroth2015distributed}
Deisenroth, Marc~Peter and Ng, Jun~Wei.
\newblock Distributed {G}aussian processes.
\newblock In \emph{International Conference on Machine Learning}, pp.\
  1481--1490. PMLR, 2015.

\bibitem[Fu et~al.(2013)Fu, Zhu, and Li]{fu2013survey}
Fu, Yifan, Zhu, Xingquan, and Li, Bin.
\newblock A survey on instance selection for active learning.
\newblock \emph{Knowledge and Information Systems}, 35\penalty0 (2):\penalty0
  249--283, 2013.

\bibitem[Gal et~al.(2014)Gal, van~der Wilk, and Rasmussen]{gal2014distributed}
Gal, Yarin, van~der Wilk, Mark, and Rasmussen, Carl~Edward.
\newblock Distributed variational inference in sparse {G}aussian process
  regression and latent variable models.
\newblock In \emph{Advances in Neural Information Processing Systems}, pp.\
  3257--3265. Curran Associates, Inc., 2014.

\bibitem[Genest \& Zidek(1986)Genest and Zidek]{genest1986combining}
Genest, Christian and Zidek, James~V.
\newblock Combining probability distributions: {A} critique and an annotated
  bibliography.
\newblock \emph{Statistical Science}, 1\penalty0 (1):\penalty0 114--135, 1986.

\bibitem[Hensman et~al.(2013)Hensman, Fusi, and Lawrence]{hensman2013gaussian}
Hensman, James, Fusi, Nicol{\`o}, and Lawrence, Neil~D.
\newblock Gaussian processes for big data.
\newblock In \emph{Proceedings of the 29th Conference on Uncertainty in
  Artificial Intelligence}, pp.\  282--290. AUAI Press, 2013.

\bibitem[Hinton(2002)]{hinton2002training}
Hinton, Geoffrey~E.
\newblock Training products of experts by minimizing contrastive divergence.
\newblock \emph{Neural Computation}, 14\penalty0 (8):\penalty0 1771--1800,
  2002.

\bibitem[Hoang et~al.(2016)Hoang, Hoang, and Low]{hoang2016distributed}
Hoang, Trong~Nghia, Hoang, Quang~Minh, and Low, Bryan Kian~Hsiang.
\newblock A distributed variational inference framework for unifying parallel
  sparse {G}aussian process regression models.
\newblock In \emph{International Conference on Machine Learning}, pp.\
  382--391. PMLR, 2016.

\bibitem[Ionescu(2015)]{ionescu2015revisiting}
Ionescu, Radu~Cristian.
\newblock Revisiting large scale distributed machine learning.
\newblock \emph{arXiv preprint arXiv:1507.01461}, 2015.

\bibitem[Lawrence(2005)]{lawrence2005probabilistic}
Lawrence, Neil.
\newblock Probabilistic non-linear principal component analysis with {G}aussian
  process latent variable models.
\newblock \emph{Journal of Machine Learning Research}, 6\penalty0
  (Nov):\penalty0 1783--1816, 2005.

\bibitem[Liu et~al.(2017)Liu, Cai, and Ong]{liu2017adaptive}
Liu, Haitao, Cai, Jianfei, and Ong, Yew-Soon.
\newblock An adaptive sampling approach for {K}riging metamodeling by
  maximizing expected prediction error.
\newblock \emph{Computers \& Chemical Engineering}, 106\penalty0
  (Nov):\penalty0 171--182, 2017.

\bibitem[Liu et~al.(2018)Liu, Cai, and Ong]{liu2018remarks}
Liu, Haitao, Cai, Jianfei, and Ong, Yew-Soon.
\newblock Remarks on multi-output {G}aussian process regression.
\newblock \emph{Knowledge-Based Systems}, 144\penalty0 (March):\penalty0
  102--121, 2018.

\bibitem[Moore \& Russell(2015)Moore and Russell]{moore2015gaussian}
Moore, David and Russell, Stuart~J.
\newblock Gaussian process random fields.
\newblock In \emph{Advances in Neural Information Processing Systems}, pp.\
  3357--3365. Curran Associates, Inc., 2015.

\bibitem[Park et~al.(2011)Park, Huang, and Ding]{park2011domain}
Park, Chiwoo, Huang, Jianhua~Z, and Ding, Yu.
\newblock Domain decomposition approach for fast {G}aussian process regression
  of large spatial data sets.
\newblock \emph{Journal of Machine Learning Research}, 12\penalty0
  (May):\penalty0 1697--1728, 2011.

\bibitem[Peng et~al.(2017)Peng, Zhe, Zhang, and Qi]{peng2017asynchronous}
Peng, Hao, Zhe, Shandian, Zhang, Xiao, and Qi, Yuan.
\newblock Asynchronous distributed variational {G}aussian process for
  regression.
\newblock In \emph{International Conference on Machine Learning}, pp.\
  2788--2797. PMLR, 2017.

\bibitem[Qui{\~n}onero-Candela \& Rasmussen(2005)Qui{\~n}onero-Candela and
  Rasmussen]{quinonero2005unifying}
Qui{\~n}onero-Candela, Joaquin and Rasmussen, Carl~Edward.
\newblock A unifying view of sparse approximate {G}aussian process regression.
\newblock \emph{Journal of Machine Learning Research}, 6\penalty0
  (Dec):\penalty0 1939--1959, 2005.

\bibitem[Ranjan \& Gneiting(2010)Ranjan and Gneiting]{ranjan2010combining}
Ranjan, Roopesh and Gneiting, Tilmann.
\newblock Combining probability forecasts.
\newblock \emph{Journal of the Royal Statistical Society: Series B (Statistical
  Methodology)}, 72\penalty0 (1):\penalty0 71--91, 2010.

\bibitem[Rasmussen \& Ghahramani(2002)Rasmussen and
  Ghahramani]{rasmussen2002infinite}
Rasmussen, Carl~E and Ghahramani, Zoubin.
\newblock Infinite mixtures of {G}aussian process experts.
\newblock In \emph{Advances in Neural Information Processing Systems}, pp.\
  881--888. Curran Associates, Inc., 2002.

\bibitem[Rasmussen \& Williams(2006)Rasmussen and
  Williams]{rasmussen2006gaussian}
Rasmussen, Carl~Edward and Williams, Christopher K.~I.
\newblock \emph{Gaussian processes for machine learning}.
\newblock MIT Press, 2006.

\bibitem[Rulli{\`e}re et~al.(2017)Rulli{\`e}re, Durrande, Bachoc, and
  Chevalier]{rulliere2017nested}
Rulli{\`e}re, Didier, Durrande, Nicolas, Bachoc, Fran{\c{c}}ois, and Chevalier,
  Cl{\'e}ment.
\newblock Nested {K}riging predictions for datasets with a large number of
  observations.
\newblock \emph{Statistics and Computing}, pp.\  1--19, 2017.

\bibitem[Seeger et~al.(2003)Seeger, Williams, and Lawrence]{seeger2003fast}
Seeger, Matthias, Williams, Christopher, and Lawrence, Neil.
\newblock Fast forward selection to speed up sparse {G}aussian process
  regression.
\newblock In \emph{Artificial Intelligence and Statistics}, pp.\
  EPFL--CONF--161318. PMLR, 2003.

\bibitem[Shahriari et~al.(2016)Shahriari, Swersky, Wang, Adams, and
  de~Freitas]{shahriari2016taking}
Shahriari, Bobak, Swersky, Kevin, Wang, Ziyu, Adams, Ryan~P, and de~Freitas,
  Nando.
\newblock Taking the human out of the loop: {A} review of {B}ayesian
  optimization.
\newblock \emph{Proceedings of the IEEE}, 104\penalty0 (1):\penalty0 148--175,
  2016.

\bibitem[Snelson \& Ghahramani(2006)Snelson and Ghahramani]{snelson2006sparse}
Snelson, Edward and Ghahramani, Zoubin.
\newblock Sparse {G}aussian processes using pseudo-inputs.
\newblock In \emph{Advances in Neural Information Processing Systems}, pp.\
  1257--1264. MIT Press, 2006.

\bibitem[Snelson \& Ghahramani(2007)Snelson and Ghahramani]{snelson2007local}
Snelson, Edward and Ghahramani, Zoubin.
\newblock Local and global sparse {G}aussian process approximations.
\newblock In \emph{Artificial Intelligence and Statistics}, pp.\  524--531.
  PMLR, 2007.

\bibitem[Tavassolipour et~al.(2017)Tavassolipour, Motahari, and
  Shalmani]{tavassolipour2017learning}
Tavassolipour, Mostafa, Motahari, Seyed~Abolfazl, and Shalmani,
  Mohammad-Taghi~Manzuri.
\newblock Learning of {G}aussian processes in distributed and communication
  limited systems.
\newblock \emph{arXiv preprint arXiv:1705.02627}, 2017.

\bibitem[Titsias(2009)]{titsias2009variational}
Titsias, Michalis~K.
\newblock Variational learning of inducing variables in sparse {G}aussian
  processes.
\newblock In \emph{Artificial Intelligence and Statistics}, pp.\  567--574.
  PMLR, 2009.

\bibitem[Tresp(2000)]{tresp2000bayesian}
Tresp, Volker.
\newblock A {B}ayesian committee machine.
\newblock \emph{Neural Computation}, 12\penalty0 (11):\penalty0 2719--2741,
  2000.

\bibitem[Vazquez \& Bect(2010)Vazquez and Bect]{vazquez2010pointwise}
Vazquez, Emmanuel and Bect, Julien.
\newblock Pointwise consistency of the {K}riging predictor with known mean and
  covariance functions.
\newblock In \emph{9th International Workshop in Model-Oriented Design and
  Analysis}, pp.\  221--228. Springer, 2010.

\bibitem[Wilson \& Nickisch(2015)Wilson and Nickisch]{wilson2015kernel}
Wilson, Andrew and Nickisch, Hannes.
\newblock Kernel interpolation for scalable structured {G}aussian processes
  ({KISS-GP}).
\newblock In \emph{International Conference on Machine Learning}, pp.\
  1775--1784. PMLR, 2015.

\bibitem[Wilson et~al.(2016)Wilson, Hu, Salakhutdinov, and
  Xing]{wilson2016deep}
Wilson, Andrew~Gordon, Hu, Zhiting, Salakhutdinov, Ruslan, and Xing, Eric~P.
\newblock Deep kernel learning.
\newblock In \emph{Artificial Intelligence and Statistics}, pp.\  370--378.
  PMLR, 2016.

\bibitem[Yuan \& Neubauer(2009)Yuan and Neubauer]{yuan2009variational}
Yuan, Chao and Neubauer, Claus.
\newblock Variational mixture of {G}aussian process experts.
\newblock In \emph{Advances in Neural Information Processing Systems}, pp.\
  1897--1904. Curran Associates, Inc., 2009.

\end{thebibliography}
	\bibliographystyle{icml2018}

%%%%%%%%%%%%%%%%%%%%%%%%%%%%%%%%%%%%%%%%%%%%%%%%%%%%%%%%%%%%%%%%%%%%%%%%%%%%%%%
%%%%%%%%%%%%%%%%%%%%%%%%%%%%%%%%%%%%%%%%%%%%%%%%%%%%%%%%%%%%%%%%%%%%%%%%%%%%%%%
% DELETE THIS PART. DO NOT PLACE CONTENT AFTER THE REFERENCES!
%%%%%%%%%%%%%%%%%%%%%%%%%%%%%%%%%%%%%%%%%%%%%%%%%%%%%%%%%%%%%%%%%%%%%%%%%%%%%%%
%%%%%%%%%%%%%%%%%%%%%%%%%%%%%%%%%%%%%%%%%%%%%%%%%%%%%%%%%%%%%%%%%%%%%%%%%%%%%%%

\appendix
\section{Proof of Proposition \ref{Prop_Typical_disjoint}} \label{App_Prop_Typical_disjoint}
%\begin{proof}
With disjoint partition, we consider two extreme local GP experts. For the first extreme expert $\mathcal{M}_{a_n}$ ($1 \le a_n \le M_n$), the test point $\bm{x}_*$ falls into the local region defined by $\bm{X}_{a_n}$, i.e., $\bm{x}_*$ is adherent to $\bm{X}_{a_n}$ when $n \to \infty$. Hence, we have \cite{vazquez2010pointwise}
\begin{equation*}
	\lim_{n \to \infty} \sigma^{2}_{a_n}(\bm{x}_*) = \lim_{n \to \infty} \sigma^{2}_{\epsilon,n} = \sigma^{2}_{\eta}.
\end{equation*}
For the other extreme expert $\mathcal{M}_{b_n}$, it lies farthest away from $\bm{x}_*$ such that the related prediction variance $\sigma^{2}_{b_n}(\bm{x}_*)$ is closest to $\sigma^{2}_{**}$. It is known that for any $\mathcal{M}_i$ ($i \ne a_n$) where $\bm{x}_*$ is away from the training data $\bm{X}_i$, given the relative distance $r_i = \min \|\bm{x}_*-\bm{x}\|_{\forall \bm{x} \in \bm{X}_{i}}$, we have $\lim_{r_i \to \infty} \sigma^{2}_{i}(\bm{x}_*) = \sigma^{2}_{**}$. Since, however, we here focus on the GP predictions in the bounded region $\Omega \in [0,1]^d$ and employ the covariance function $k(.) > 0$, then the positive sequence $c_n = \{\sigma^{-2}_{b_n}(\bm{x}_*) - \sigma^{-2}_{**} \}$ is small but satisfies $\lim_{n \to \infty} c_n > 0$ and
\begin{equation*}
\sigma^{-2}_{i}(\bm{x}_*) - \sigma^{-2}_{**} \ge c_n, \, 1 \le i \ne a_n \le M_n.
\end{equation*}
The equality holds only when $i = b_n$.

Thereafter, with the sequence $\epsilon_n=\min\{c_n,\frac{1}{M_n^{\alpha}}\} \to_{n \to \infty} 0$ where $\alpha > 0$ we have
\begin{equation*}
\sigma^{-2}_{i}(\bm{x}_*) - \sigma^{-2}_{**} \ge c_n \ge \epsilon_n, \, 1 \le i \ne a_n \le M_n.
\end{equation*} 
It is found that $c_n = \epsilon_n$ is possible to hold only when $M_n$ is small. With the increase of $n$, $\epsilon_n$ quickly becomes much smaller than $c_n$ since $\lim_{n \to \infty} 1/M_n^{\alpha} = 0$.

The typical aggregated prediction variance writes
\begin{equation} \label{Eq_disjoint_s2}
\begin{aligned}    
\sigma^{-2}_{\mathcal{A},n}(\bm{x}_*) & = \sum_{i=1}^{M_n} \beta_{i} (\sigma^{-2}_{i}(\bm{x}_*)-\sigma^{-2}_{**}) + \sigma^{-2}_{**},
\end{aligned}
\end{equation}
where for (G)PoE we remove the prior precision $\sigma^{-2}_{**}$. We prove below the inconsistency of (G)PoE and (R)BCM using disjoint partition.

For PoE, \eqref{Eq_disjoint_s2} is $\sum_{i=1}^{M_n}\sigma_{i}^{-2}(\bm{x}_*) > M_n \sigma^{-2}_{**} \to_{n \to \infty} \infty$, leading to the inconsistent variance $\lim_{n \to \infty} \sigma^{2}_{\mathcal{A},n} = 0$. For (R)BCM, the first term of $\sigma^{-2}_{\mathcal{A},n}(\bm{x}_*)$ in \eqref{Eq_disjoint_s2} satisfies, given that $n$ is large enough,
\begin{equation*}
\sum_{i=1}^{M_n} \beta_{i} (\sigma^{-2}_{i}(\bm{x}_*)-\sigma^{-2}_{**}) > \epsilon_n \sum_{i=1}^{M_n} \beta_{i} = \frac{1}{M_n^{\alpha}} \sum_{i=1}^{M_n} \beta_{i}.
\end{equation*}
Taking $\beta_{i} = 1$ for BCM and $\alpha = 0.5$, we have $\frac{1}{M_n^{\alpha}} \sum_{i=1}^{M_n} \beta_{i} = \sqrt{M_n} \to_{n \to \infty} \infty$, leading to the inconsistent variance $\lim_{n \to \infty} \sigma^{2}_{\mathcal{A},n} = 0$. For RBCM, since 
\begin{equation*}
\beta_{i} = 0.5(\log\sigma_{**}^{2} - \log\sigma_{i}^2(\bm{x}_*)) \ge 0.5\log(1+c_n\sigma^{2}_{**})
\end{equation*}
where the equality holds only when $i = b_n$, we have $\frac{1}{M_n^{\alpha}} \sum_{i=1}^{M_n} \beta_{i} > 0.5\log(1+c_n\sigma^{2}_{**}) \sqrt{M_n} \to_{n \to \infty} \infty$, leading to the inconsistent variance $\lim_{n \to \infty} \sigma^{2}_{\mathcal{A},n} = 0$. 

Finally, for GPoE, we know that when $n \to \infty$, $\sigma^{-2}_{a_n}(\bm{x}_*)$ converges to $\sigma^{-2}_{\eta}$; but the other prediction precisions satisfy $c_n + \sigma^{-2}_{**} \le \sigma^{-2}_i(\bm{x}_*) < \sigma^{-2}_{\epsilon,n} \to_{n \to \infty} \sigma_{\eta}^{-2}$ for $1 \le i \ne a_n \le M_n$, since $\bm{x}_*$ is away from their training points. Hence, we have
\begin{equation*}
\begin{aligned}
	& \lim_{n \to \infty} \left(\sigma^{-2}_{\eta} - \sigma^{-2}_{\mathrm{GPoE}}(\bm{x}_*)\right) \\
	=& \lim_{n \to \infty} \frac{1}{M_n} \left( \sigma^{-2}_{\eta} - \sigma^{-2}_{a_n}(\bm{x}_*) \right) \\
	&+ \lim_{n \to \infty} \frac{1}{M_n} \sum_{i \ne a_n}^{M_n} \left( \sigma^{-2}_{\eta} - \sigma^{-2}_{i}(\bm{x}_*) \right) \\
	>& \lim_{n \to \infty} \frac{1}{M_n} \left( \sigma^{-2}_{\eta} - \sigma^{-2}_{a_n}(\bm{x}_*) \right) \\
	&+ \lim_{n \to \infty} \frac{1}{M_n} \sum_{i \ne a_n}^{M_n} \left( \sigma^{-2}_{\eta} - \sigma^{-2}_{\epsilon,n}(\bm{x}_*) \right) = 0,
\end{aligned}
\end{equation*}
which means that $\sigma^{2}_{\mathrm{GPoE}}(\bm{x}_*)$ is inconsistent since $\lim_{n \to \infty} \sigma^{2}_{\mathrm{GPoE}}(\bm{x}_*) > \sigma^{2}_{\eta}$. Meanwhile, we easily find that $\lim_{n \to \infty} \sigma^{-2}_{\mathrm{GPoE}}(\bm{x}_*) > c_n + \sigma^{-2}_{**}$, leading to $\lim_{n \to \infty} \sigma^{2}_{\mathrm{GPoE}}(\bm{x}_*) < \sigma^2_{b_n}(\bm{x}_*) < \sigma^{2}_{**}$.	%\qedhere
%\end{proof}

\section{Proof of Proposition \ref{Prop_Typical_random}} \label{App_Prop_Typical_random}
%\begin{proof}
With smoothness assumption and particularly distributed noise (normal or Laplacian distribution), it has been proved that the GP predictions would converge to the true predictions when $n \to \infty$ \cite{choi2004posterior}. Hence, given that the points in $\bm{X}_{i}$ are randomly selected without replacement from $\bm{X}$ and $n_i = n/M_n \to_{n \to \infty} \infty$,  we have 
\begin{equation*}
\lim_{n \to \infty} \mu_{i}(\bm{x}_*) = \mu_{\eta}(\bm{x}_*), \lim_{n\to \infty} \sigma^2_{i}(\bm{x}_*) = \sigma^2_{\eta}, \quad 1 \le i \le M_n.
\end{equation*}

For the aggregated prediction variance, we have
\begin{equation*}
\lim_{n \to \infty} \sigma_{\mathcal{A},n}^{-2}(\bm{x}_*) = \lim_{n \to \infty} \left[ \sum_{i=1}^{M_n} \beta_{i} (\sigma_{i}^{-2}(\bm{x}_*) -\sigma_{**}^{-2}) + \sigma_{**}^{-2} \right], 
\end{equation*}
where for (G)PoE we remove $\sigma^{-2}_{**}$. For PoE, given $\beta_i=1$ and $\lim_{n \to \infty} \sigma^{-2}_{i}(\bm{x}_*) = \sigma^{-2}_{\eta}$, we have the inconsistent variance $\lim_{n \to \infty} \sigma_{\mathcal{A},n}^{-2}(\bm{x}_*) = \lim_{n \to \infty} M_n \sigma_{\eta}^{-2} = \infty$. For GPoE, given $\beta_{i}=1/M_n$ we have the consistent variance $\lim_{n \to \infty} \sigma_{\mathcal{A},n}^{-2}(\bm{x}_*) = M_n \frac{1}{M_n} \sigma_{\eta}^{-2} = \sigma_{\eta}^{-2}$. For BCM, given $\beta_i=1$ we have  the inconsistent variance $\lim_{n \to \infty} \sigma_{\mathcal{A},n}^{-2}(\bm{x}_*) = \lim_{n \to \infty} [M_n (\sigma_{\eta}^{-2} -\sigma_{**}^{-2}) + \sigma_{**}^{-2}] = \infty$. Finally, for RBCM, given  $\lim_{n \to \infty} \beta_{i} = \overline{\beta} = 0.5\log(\sigma^2_{**}/\sigma^2_{\eta})$, we have the inconsistent variance $\lim_{n \to \infty} \sigma_{\mathcal{A},n}^{-2}(\bm{x}_*) = \lim_{n \to \infty} [M_n \overline{\beta} (\sigma_{\eta}^{-2} -\sigma_{**}^{-2}) + \sigma_{**}^{-2}] = \infty$.

Then, for the aggregated prediction mean we have
\begin{equation*}
\lim_{n \to \infty} \mu_{\mathcal{A},n}(\bm{x}_*) =  \lim_{n \to \infty} \sigma_{\mathcal{A},n}^{2}(\bm{x}_*) \sum_{i=1}^{M_n} \beta_i \sigma^{-2}_{i}(\bm{x}_*) \mu_{i}(\bm{x}_*).
\end{equation*}
For PoE, given $\beta_i = 1$ and $\lim_{n \to \infty} \sigma^{-2}_{i}(\bm{x}_*) / \sigma_{\mathcal{A},n}^{-2}(\bm{x}_*) = 1/M_n$, we have the consistent prediction mean $\lim_{n \to \infty} \mu_{\mathcal{A},n}(\bm{x}_*) = \mu_{\eta}(\bm{x}_*)$. For GPoE, given $\beta_i = 1/M_n$ and $\lim_{n \to \infty} \sigma^{-2}_{i}(\bm{x}_*) / \sigma_{\mathcal{A},n}^{-2}(\bm{x}_*) = 1$, we have the consistent prediction mean $\lim_{n \to \infty} \mu_{\mathcal{A},n}(\bm{x}_*) = \mu_{\eta}(\bm{x}_*)$. For (R)BCM, given $\beta_i = \overline{\beta} = 1$ or $\lim_{n \to \infty} \beta_i = \overline{\beta} = 0.5\log(\sigma^2_{**}/\sigma^2_{\eta})$, we have the inconsistent prediction mean $\lim_{n \to \infty} \mu_{\mathcal{A},n}(\bm{x}_*) = \lim_{n \to \infty} \overline{\beta} \sigma^{-2}_{\eta} \mu_{\eta}(\bm{x}_*) / (\overline{\beta} (\sigma^{-2}_{\eta} - \sigma^{-2}_{**}) +  \sigma^{-2}_{**}/M_n)=  a \mu_{\eta}(\bm{x}_*)$ where $a = \sigma_{\eta}^{-2}/(\sigma_{\eta}^{-2} - \sigma_{**}^{-2}) \geq 1$ and the equality holds when $\sigma^2_{\eta} = 0$. %\qedhere
%\end{proof}

\section{Proof of Proposition \ref{Prop_GRBCM}} \label{App_Prop_GRBCM}
%\begin{proof}
Given that the points in the communication subset $\mathcal{D}_c$ are randomly selected without replacement from $\mathcal{D}$ and $n_c = n/M_n \to_{n \to \infty} \infty$, we have $\lim_{n \to \infty} \mu_c(\bm{x}_*) = \mu_{\eta}(\bm{x}_*)$ and $\lim_{n \to \infty} \sigma^2_c(\bm{x}_*) =  \sigma^2_{\eta}$ for $\mathcal{M}_{c}$. Likewise, for the expert $\mathcal{M}_{+i}$ trained on the augmented dataset $\mathcal{D}_{+i} = \{\mathcal{D}_i, \mathcal{D}_c\}$ with size $n_{+i} = 2n/M_n$, we have $\lim_{n \to \infty} \mu_{+i}(\bm{x}_*) =  \mu_{\eta}(\bm{x}_*)$ and $\lim_{n \to \infty} \sigma^2_{+i}(\bm{x}_*) =  \sigma^2_{\eta}$ for $2 \le i \le M$. 

We first derive the upper bound of $\sigma^2_c(\bm{x}_*)$. For the stationary covariance function $k(.)>0$, when $n_c$ is large enough we have \cite{vazquez2010pointwise}
\begin{equation*}
\sigma^2_c(\bm{x}_*) \le k(\bm{x}_*,\bm{x}_*) - \frac{k^2(\bm{x}_*,\bm{x}')}{k(\bm{x}',\bm{x}')} + \sigma^2_{\epsilon,n},
\end{equation*} 
where $\bm{x}' \in \bm{X}_c$ is the nearest data point to $\bm{x}_*$. It is known that the relative distance $r_{c} = \|\bm{x}_*-\bm{x}'\|$ is proportional to the inverse of the training size $n_c$, i.e., $r_{c} \propto 1/n_c = M_n/n \to_{n \to \infty} 0$. Conventional stationary covariance functions only relay on the relative distance (once the covariance parameters have been determined) and decrease with $r_c$. Consequently, the prediction variance $\sigma^2_c(\bm{x}_*) $ increases with $r_c$.
Taking the SE covariance function in~\eqref{Eq_SE_Kernel} for example,\footnote{We take the SE kernel for example since conventional kernels, e.g., the rational quadratic kernel and the Mat\'ern class of kernels, can reduce to the SE kernel under some conditions.} when $r_c \to 0$ we have, given $l_0 = \min_{1 \le i \le d} \{l_i\}$,
\begin{equation} \label{Eq_sigma2_t_UB}
\begin{aligned}
\sigma^2_c(\bm{x}_*) &\le \sigma^2_{f}  - \sigma^2_{f} \exp(-r_c^2/l_0^2) + \sigma^2_{\epsilon,n} \\
&< \frac{\sigma^2_{f}}{l_0^2}r_c^2 + \sigma^2_{\epsilon,n} = a r_c^2 + \sigma^2_{\epsilon,n}.
\end{aligned}
\end{equation}
We clearly see from this inequality that when $r_c \to 0$, $\sigma^2_c(\bm{x}_*)$ goes to $\sigma^2_{\eta}$ since $\lim_{n \to \infty} \sigma^2_{\epsilon,n} = \sigma^2_{\eta}$.

Then, we rewrite the precision of GRBCM in \eqref{Eq_s2_GRBCM} as, given $\beta_2 = 1$,
\begin{equation} \label{Eq_newS2_GRBCM}
\sigma_{\mathrm{GRBCM}}^{-2}(\bm{x}_*) = \sigma_{+2}^{-2}(\bm{x}_*) + \sum_{i=3}^{M_n} \beta_i \left(\sigma_{+i}^{-2}(\bm{x}_*)  - \sigma_c^{-2}(\bm{x}_*) \right).
\end{equation}
Compared to $\mathcal{M}_c$, $\mathcal{M}_{+i}$ is trained on a more dense dataset $\mathcal{D}_{+i}$, leading to $\sigma^{2}_{+i}(\bm{x}_*) \le \sigma^{2}_c(\bm{x}_*)$ for a large enough $n$.\footnote{The equality is possible to hold when we employ disjoint partition for $\{\mathcal{D}_i\}_{i=2}^{M_n}$ and $\bm{x}_*$ is away from $\bm{X}_i$.} 

Given \eqref{Eq_sigma2_t_UB} and $\sigma^2_{+i}(\bm{x}_*) > \sigma^2_{\epsilon,n}$, the weight $\beta_i$ satisfies, for $3 \le i \le M_n$,
\begin{equation} \label{Eq_beta_i_UB}
\begin{aligned}
0 \le \beta_i &= \frac{1}{2} \log \left(\frac{\sigma_c^2(\bm{x}_*)}{\sigma_{+i}^2(\bm{x}_*)} \right) < \frac{1}{2} \log \left(\frac{\sigma_c^2(\bm{x}_*)}{\sigma^2_{\epsilon,n}} \right) \\
&< \frac{1}{2} \log \left(\frac{a r_c^2 + \sigma^2_{\epsilon,n}}{\sigma^2_{\epsilon,n}} \right) \le \frac{a}{2\sigma^2_{\epsilon,n}} r_c^2.
\end{aligned}
\end{equation}
Besides, the precision discrepancy satisfies, for $3 \le i \le M_n$,
\begin{equation} \label{Eq_sigma2_i_UB}
\begin{aligned}
0 \le \sigma_{+i}^{-2}(\bm{x}_*)  - \sigma_c^{-2}(\bm{x}_*) &= \sigma_c^{-2}(\bm{x}_*) \left(\frac{\sigma_c^2(\bm{x}_*)}{\sigma_{+i}^2(\bm{x}_*)} -1 \right) \\
&< \frac{1}{\sigma^2_{\epsilon,n}} \frac{a}{\sigma^2_{\epsilon,n}} r_c^2.
\end{aligned}
\end{equation}
Hence, the second term in the right-hand side of \eqref{Eq_newS2_GRBCM} satisfies
\begin{equation*}
\sum_{i=3}^{M_n} \beta_i \left(\sigma_{+i}^{-2}(\bm{x}_*)  - \sigma_c^{-2}(\bm{x}_*) \right) < \sum_{i=3}^{M_n} \frac{a^2}{2 \sigma^6_{\epsilon,n}} r_c^4 \propto \frac{M_n^5}{n^4}.
\end{equation*}
Since $\lim_{n \to \infty} n/M_n^2 > 0$, we have $\lim_{n \to \infty} n^4/M_n^5 = \infty$, and furthermore,
\begin{equation} \label{Eq_zero_GRBCM}
\lim_{n \to \infty} \sum_{i=3}^{M_n} \beta_i \left(\sigma_{+i}^{-2}(\bm{x}_*)  - \sigma_c^{-2}(\bm{x}_*) \right) = 0.
\end{equation}
Substituting \eqref{Eq_zero_GRBCM} and $\lim_{n \to \infty} \sigma_{+2}^{-2}(\bm{x}_*) = \sigma^{-2}_{\eta}$ into \eqref{Eq_newS2_GRBCM}, we have a consistent prediction precision as
\begin{equation*}
\lim_{n \to \infty} \sigma_{\mathrm{GRBCM}}^{-2}(\bm{x}_*) = \sigma^{-2}_{\eta}.
\end{equation*}

Similarly, we rewrite the GRBCM's prediction mean in \eqref{Eq_mu_GRBCM} as
\begin{equation} \label{Eq_newMu_GRBCM}
\mu_{\mathrm{GRBCM}}(\bm{x}_*) = \sigma_{\mathrm{GRBCM}}^2(\bm{x}_*) \left(\mu_{\Delta} + \sigma_{+2}^{-2}(\bm{x}_*)\mu_{+2}(\bm{x}_*) \right),
\end{equation}
where
\begin{equation*}
\mu_{\Delta} = \sum_{i=3}^{M_n} \beta_i \left(\sigma_{+i}^{-2}(\bm{x}_*) \mu_{+i}(\bm{x}_*) - \sigma_c^{-2}(\bm{x}_*)\mu_c(\bm{x}_*) \right).
\end{equation*}
Let $\delta_{\mathrm{max}} = \max_{3 \le i \le M_n} \left|\frac{\sigma_c^{2}(\bm{x}_*)}{\sigma_{+i}^{2}(\bm{x}_*)} \mu_{+i}(\bm{x}_*) - \mu_c(\bm{x}_*)\right| \to_{n \to \infty} 0$, we have
	\begin{equation} \label{Eq_zero2_GRBCM}
	\begin{aligned}
	\left|\mu_{\Delta} \right| &\le \sum_{i=3}^{M_n} \beta_i \sigma_c^{-2} \left|\frac{\sigma_c^{2}(\bm{x}_*)}{\sigma_{+i}^{2}(\bm{x}_*)} \mu_{+i}(\bm{x}_*) - \mu_c(\bm{x}_*)\right| \\
	&\mathop{<}^{\mathrm{Eq.} \eqref{Eq_beta_i_UB}} \sum_{i=3}^{M_n} \frac{a r_c^2}{2\sigma^4_{\epsilon,n}} \delta_{\mathrm{max}} \to_{n \to \infty} 0.
	\end{aligned}
	\end{equation}   
Substituting \eqref{Eq_zero2_GRBCM} into \eqref{Eq_newMu_GRBCM}, we have the consistent prediction mean as
\begin{equation*}
\lim_{n \to \infty} \mu_{\mathrm{GRBCM}}(\bm{x}_*) = \mu_{\eta}(\bm{x}_*). %\qedhere
\end{equation*} 
%\end{proof}

\section{Discussions of GRBCM on the toy example} \label{App_GRBCM_toy}
It is observed that the proposed GRBCM showcases superiority over existing aggregations on the toy example, which is brought by the particularly designed aggregation structure: the global communication expert $\mathcal{M}_c$ to capture the long-term features of the target function, and the remaining experts $\{\mathcal{M}_{+i}\}_{i=2}^M$ to refine local predictions.

\begin{figure}[ht]
	\vskip 0.0in
	\begin{center}
		\centerline{\includegraphics[width=0.9\columnwidth]{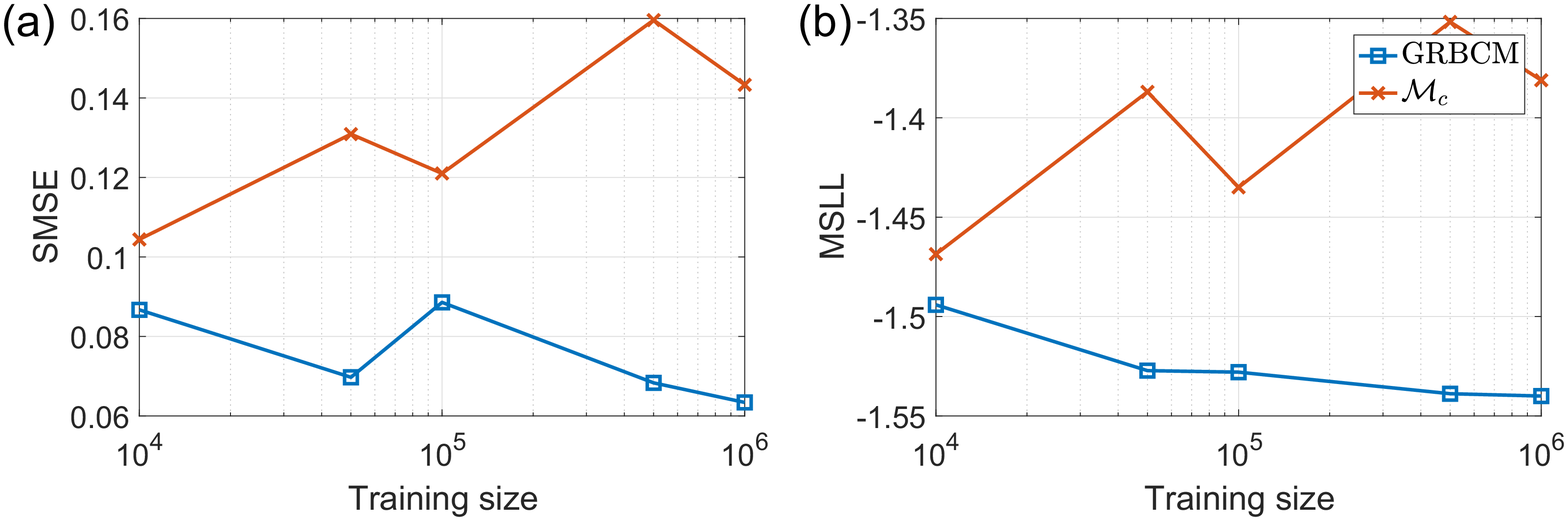}}
		\caption{Comparative results of GRBCM and $\mathcal{M}_c$ on the toy example.}
		\label{Fig_Results_GRBCM_Mc}
	\end{center}
	\vskip -0.3in
\end{figure}

To verify the capability of GRBCM, we compare it with the pure global expert $\mathcal{M}_c$ which relies on a random subset $\bm{X}_c$. Fig.~\ref{Fig_Results_GRBCM_Mc} shows the comparative results of GRBCM and $\mathcal{M}_c$ on the toy example. It is found that with increasing $n$, (i) GRBCM always outperforms $\mathcal{M}_c$ because of the benefits brought by local experts; and (ii) the predictions of $\mathcal{M}_c$ generally become poorer since it becomes intractable to choose a good subset from the increasing dataset.

\section{Experimental results of NPAE} \label{App_results_NPAE}
Table~\ref{Tab_Results_GRBCM_NPAE} compares the results of GRBCM and NPAE over 10 runs on the \textit{kin40k} dataset ($M=16$) and the \textit{sarcos} dataset ($M=72$) using disjoint partition. It is observed that GRBCM performs slightly better than NPAE on the \textit{kin40k} dataset, and produces competitive results on the \textit{sarcos} dataset. But in terms of the computing efficiency, since NPAE needs to build and invert an $M \times M$ covariance matrix at each test point, it requires much more running time, especially for the \textit{sarcos} dataset with $M = 72$.

\begin{table}[t]
	\caption{Comparative results (mean and standard deviation) of GRBCM and NPAE over 10 runs on the \textit{kin40k} dataset ($M=16$) and the \textit{sarcos} dataset ($M=72$) using disjoint partition. The computing time $t$ for each model involves the training and predicting time.}
	\label{Tab_Results_GRBCM_NPAE}
	\vskip -0.1in
	\begin{center}
		\begin{small}
			\begin{sc}
				\begin{tabular}{lcc}
					\toprule
					\textit{kin40k} & GRBCM & NPAE\\
					\midrule
					SMSE & 0.0223 $\pm$ 0.0005 & 0.0246 $\pm$ 0.0007 \\
					MSLL & -1.9927 $\pm$ 0.0177 & -1.9565 $\pm$ 0.0170 \\
					$t$ [s]& 78.1 $\pm$ 4.4 & 2852.4 $\pm$ 16.7 \\
					\toprule
					\textit{sarcos} & GRBCM & NPAE\\
					\midrule
					SMSE  & 0.0074 $\pm$ 0.0002 & 0.0054 $\pm$ 0.0001  \\
					MSLL     & -2.3681 $\pm$ 0.0242 & -2.5900 $\pm$ 0.0068  \\
					$t$ [s]& 445.6 $\pm$ 49.4 & 26444.0 $\pm$ 1213.0 \\
					\bottomrule
				\end{tabular}
			\end{sc}
		\end{small}
	\end{center}
	\vskip -0.1in
\end{table}

%%%%%%%%%%%%%%%%%%%%%%%%%%%%%%%%%%%%%%%%%%%%%%%%%%%%%%%%%%%%%%%%%%%%%%%%%%%%%%%
%%%%%%%%%%%%%%%%%%%%%%%%%%%%%%%%%%%%%%%%%%%%%%%%%%%%%%%%%%%%%%%%%%%%%%%%%%%%%%%

\end{document}